%% file: main.tex
\pdfoutput=1

\documentclass[11pt]{article}

\usepackage[preprint]{acl}

\usepackage{times}
\usepackage{latexsym}

\usepackage[T1]{fontenc}

\usepackage[utf8]{inputenc}

\usepackage{microtype}

\usepackage{graphicx}
\usepackage{subcaption}
\usepackage{amsfonts}
\usepackage{caption}
\usepackage{colortbl}
\usepackage{color}
\usepackage[most]{tcolorbox} 
\tcbuselibrary{breakable}
\usepackage{multirow}
\usepackage{multicol}
\usepackage{adjustbox}
\usepackage{tabularx}
\usepackage{enumitem}
\usepackage{booktabs}
\usepackage{array}
\usepackage{longtable}
\usepackage{url}
\usepackage{threeparttable}

\usepackage{xcolor} 
\definecolor{brickred}{rgb}{0.8, 0.25, 0.33}
\newcommand{\redbf}[1]{\textbf{\textcolor{brickred}{#1}}}
\definecolor{purple}{HTML}{8e86be}
\definecolor{green}{HTML}{78c0ba}
\definecolor{yellow}{HTML}{e9b64e}

\title{Not All Features Deserve Attention: Graph-Guided Dependency Learning for Tabular Data Generation with Language Models}

\author{Zheyu Zhang \;  Shuo Yang \; Bardh Prenkaj \;  
         Gjergji Kasneci \vspace{0.3cm}\\
    Technical University of Munich \\
{\small \tt \{name.surname\}@tum.de}}

\begin{document}
\maketitle
\begin{abstract}
Large Language Models (LLMs) have shown strong potential for tabular data generation by modeling textualized feature-value pairs. However, tabular data inherently exhibits sparse feature-level dependencies, where many feature interactions are structurally insignificant. This creates a fundamental mismatch as LLMs' self-attention mechanism inevitably distributes focus across all pairs, diluting attention on critical relationships, particularly in datasets with complex dependencies or semantically ambiguous features. To address this limitation, we propose \textbf{GraDe} (\textbf{Gra}ph-Guided \textbf{De}pendency Learning), a novel method that explicitly integrates sparse dependency graphs into LLMs' attention mechanism. GraDe employs a lightweight dynamic graph learning module guided by externally extracted functional dependencies, prioritizing key feature interactions while suppressing irrelevant ones. Our experiments across diverse real-world datasets demonstrate that GraDe outperforms existing LLM-based approaches by up to 12\% on complex datasets while achieving competitive results with state-of-the-art approaches in synthetic data quality. Our method is minimally intrusive yet effective, offering a practical solution for structure-aware tabular data modeling with LLMs.
\footnote{Our code is available at \url{https://github.com/oooranz/GraDe}.}
\end{abstract}

\section{Introduction}
Tabular data forms the foundations of countless real-world applications across healthcare analytics \citep{fatima2017survey}, financial modeling \citep{DASTILE2020106263}, demographic research \citep{NEURIPS2021_32e54441}, and scientific experimentation \citep{10.1016/j.inffus.2021.11.011}. Generating high-quality synthetic tabular data has become increasingly important for data augmentation, privacy preservation, and model testing \citep{pmlr-v235-van-breugel24a}. However, this task presents unique challenges due to the complex structural properties inherent to tabular datasets.

\begin{figure}[t!]
    \centering
    \includegraphics[width=1\linewidth]{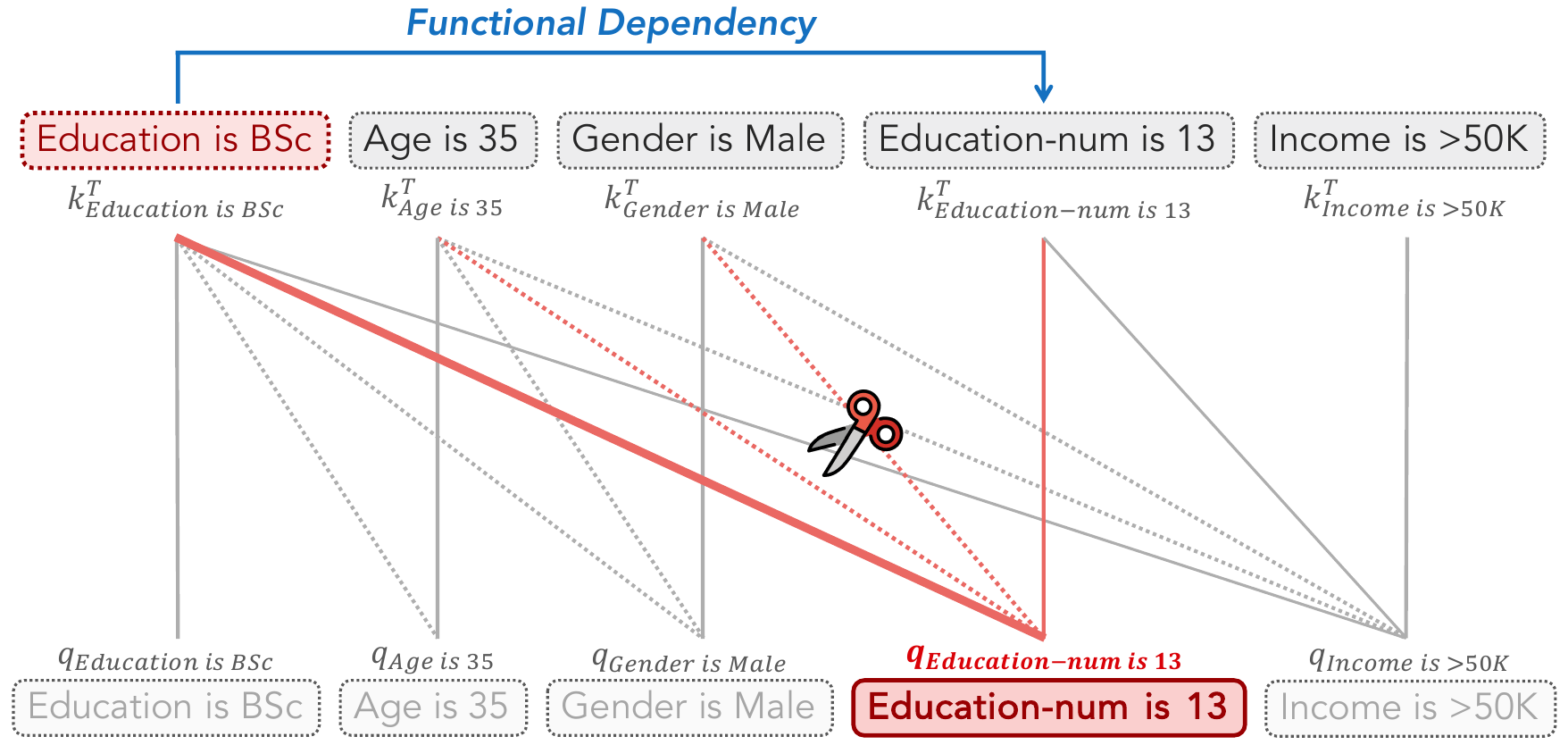}
    \caption{Structural mismatch between LLMs' dense attention patterns and tabular data's sparse dependency structure.}
    \label{fig:mismatch}
    \vspace{-10pt}
\end{figure}

Unlike images or text, tabular data exhibits distinct patterns of relationships between features. Some features have strong correlations, while others remain entirely independent from one another \citep{wang2024challenges}. More critically, certain features may uniquely determine the values of others through logical or business rules. Database theory formalizes these patterns as \textit{Functional Dependencies (FDs)}, where knowing the value of one feature (or a set of features) allows us to determine another's with certainty \citep{liu2010discover}. For example, postal codes uniquely determine cities and states, while having no connection to personal attributes like age or income. Without preserving these dependencies, synthetic data suffers from logical inconsistencies such as customer IDs linked to multiple conflicting personal details, rendering the generated data unrealistic for practical applications.

Recent years have seen rapid progress in tabular data generation \citep{xu2019modeling, liu2023goggle, kotelnikov2023tabddpm, zhang2023mixed}. Among these, Large Language Models (LLMs) offer a compelling solution due to their strong generalization capabilities from pre-training on vast corpora and understanding of complex real-world relationships \citep{liu2023pre, zhang2023baby}. \citet{borisov2022language} first applied LLMs to this task by converting data into text sequences (e.g., ``\textit{Age is 39, Income is $\leq$50K}'') and randomly shuffling feature orders during training, allowing the model to generate data conditioned on any subset of features. \citet{xu2024llms} later showed that this random shuffling disrupts the model's ability to capture dependencies between features, proposing a fixed topological ordering approach instead. This reveals a fundamental design challenge: how to maintain both flexible conditioning and accurate modeling of feature relationships in LLM-based approaches.

The root cause of this challenge is a \textit{structural mismatch} between LLMs and tabular data. As shown in Figure \ref{fig:mismatch}, LLMs use attention mechanisms where every token potentially relates to every other token \citep{vaswani2017attention}. In contrast, tabular data has sparse, non-sequential dependencies where most features are conditionally independent. This fundamental disconnect creates a dilemma: random feature ordering offers flexibility but weakens dependency modeling, while fixed ordering better captures relationships but sacrifices flexibility. When feature-value pairs are linearized as text, LLMs face the inherently difficult task of correctly interpreting each pair while preventing interference from unrelated features \citep{yan2024making}. To fully leverage LLMs for tabular data, we need an approach that maintains flexible ordering while explicitly modeling the sparse dependency structure inherent in tabular data.

To bridge this structural gap, we propose \textbf{GraDe} (\textbf{Gra}ph-Guided \textbf{De}pendency Learning). Unlike previous approaches that rely on implicit learning through feature ordering, GraDe explicitly enhances LLMs by incorporating learnable sparse dependency graphs into the attention mechanism (\S\ref{subsec:gynamic_graph_attn}). Our method dynamically models token-level relationships while leveraging externally extracted functional dependencies as guidance (\S\ref{subsec:incorporating_fd}), allowing focused attention on structurally important connections. The model is trained with a multi-objective loss balancing language fluency, structural sparsity, and functional dependency alignment (§\ref{subsec:fine_tuning}). For parameter efficiency, we also introduce \textbf{GraDe-Light}, a variant that only updates our attention modules while maintaining competitive performance. Overall, our \textbf{contributions} are as follows:

\begin{itemize}[leftmargin=*, nolistsep]
\vspace{5pt}
\item We formalize the structural mismatch problem between LLMs' dense attention patterns and tabular data's sparse dependency structure, providing a framework for understanding current methodological limitations (\S\ref{sec: Problem Formulation}).
\vspace{5pt}
\item We develop a graph-guided attention mechanism that dynamically models sparse dependencies between tokens and integrates functional dependency knowledge as soft supervision, enhancing structural awareness while preserving the underlying LLM architecture (\S\ref{sec: Methodology}).
\vspace{5pt}
\item We empirically demonstrate that both GraDe and GraDe-Light show significant improvements over existing LLM-based approaches while achieving competitive results with state-of-the-art methods. Our approach performs particularly well on datasets with complex dependency structures and maintains advantages in low-resource settings (\S\ref{sec:results}).
\vspace{5pt}
\end{itemize}

\section{Related Work}
\paragraph{Deep Generative Models for Tabular Modeling}
Early research on tabular data generation adapted deep generative models such as GANs, VAEs, and diffusion models \citep{10.1016/j.inffus.2021.11.011}. CTGAN \citep{xu2019modeling} and CTAB-GAN \citep{zhao2021ctab} tackled mixed-type data through conditional sampling and specialized normalization strategies \citep{grinsztajn2022tree}, while TVAE \citep{xu2019modeling} employed a variational framework for more stable training. More recent diffusion-based methods like TabDDPM \citep{kotelnikov2023tabddpm} and TabSyn \citep{zhang2023mixed} improved sample quality through iterative denoising of multimodal distributions. Notably, GOGGLE \citep{liu2023goggle} introduced explicit graph-based modeling of feature dependencies. Despite these advances, most generative approaches still struggle to flexibly capture the sparse dependency structures inherent in tabular data.

\paragraph{Language Models for Tabular Modeling}
Another line of research models tabular data by serializing rows into textual sequences and applying large language models for generation. GReaT \citep{borisov2022language} pioneered this direction by serializing feature-value pairs as text and employing random feature permutations to enhance flexibility. However, this randomization weakens the model's ability to learn structural dependencies \citep{muller2023attending}. Later works such as TabLLM \citep{hegselmann2023tabllm} and TAPTAP \citep{zhang2023generative} proposed refined serialization formats and large-scale tabular pretraining to better preserve relational patterns. Recent work such as P-TA \citep{yang2024p} further explored enhancing LLM-based tabular generation through policy optimization, while SPADA \citep{yang2025doubling} uses LLMs to build sparse feature dependencies in tabular data. These efforts highlight the need for stronger structural guidance beyond text serialization alone.

\paragraph{Injecting Structural Information into LMs}
The importance of aligning model architecture with data structure has been increasingly recognized across domains \citep{Kitaev2020Reformer, tay2021synthesizer, bi2024visual, abs-2502-12119}. Various approaches address this alignment: sparse attention models \citep{child2019generating, beltagy2020longformer} constrain token interactions to relevant neighborhoods; graph-enhanced methods \citep{yang2021graphformers} incorporate explicit relational structures; and specialized architectures like GraSAME \citep{yuan2024grasame} inject structural information directly into attention mechanisms. These developments reflect growing evidence that standard dense attention inadequately captures the inherent sparsity in real-world data \citep{li2022uniformer}. Our GraDe method extends this insight to tabular modeling by introducing a graph-guided attention mechanism that leverages functional dependencies to focus on meaningful feature relationships while preserving the generative flexibility of language models.

\section{Problem Formulation}
\label{sec: Problem Formulation}
\paragraph{Modeling Tabular Data as Language}
Generative modeling for tabular data aims to learn a probability distribution $p_X$ over tabular samples $\mathbf{x}\in \mathcal{X}\subseteq \mathbb{R}^{d}$ \citep{9998482}. Given a training dataset $\mathcal{D}$ consisting of $N$ i.i.d. samples $\mathbf{x}\sim p_X$, the objective is to approximate $p_X$ by learning parameters $\theta$ of a generative distribution $p_\theta$. While traditional approaches model this distribution directly in the feature space, language model-based methods typically transform tabular data into textual representations to model the joint distribution as:
\begin{equation}
p_\theta(\mathbf{x}) = \prod_{t=1}^{T} p_\theta(w_t \mid w_{<t}),
\end{equation}
where $w_{1:T} = w(\mathbf{x})$ denotes a linearized textual sequence of feature-value pairs, such as \textit{“Feature$_1$ is $x_1$, ..., Feature$_d$ is $x_d$”}. 

\begin{figure*}
    \centering
    \includegraphics[width=1\linewidth]{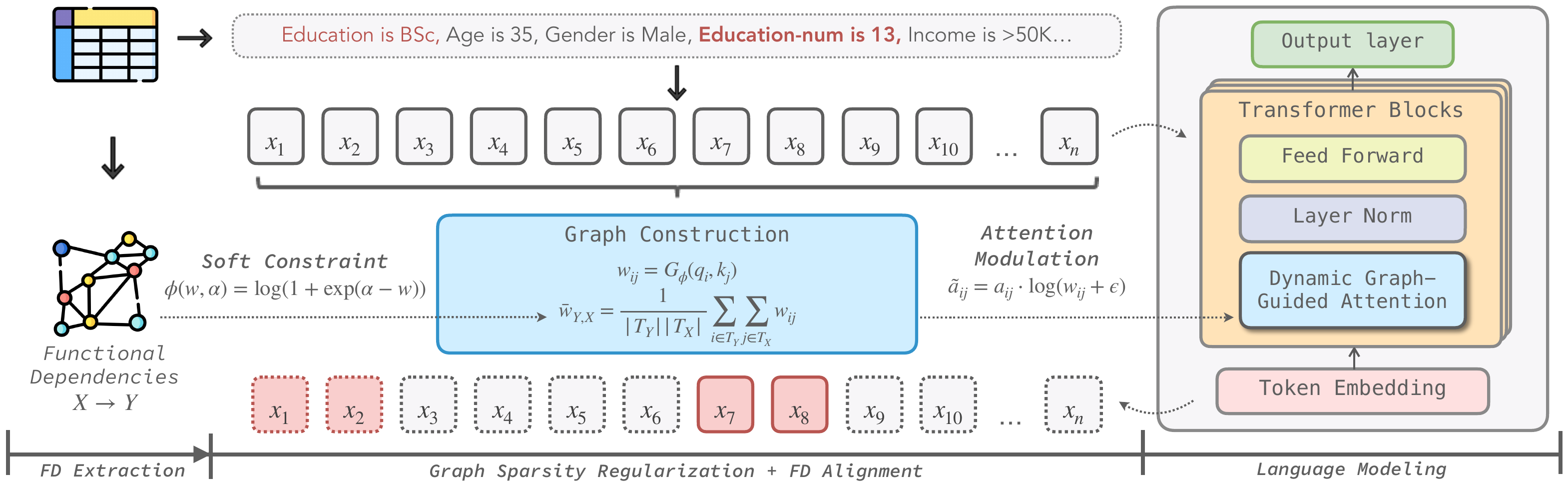}
    \caption{Pipeline of GraDe. Following \citet{borisov2022language}, we first linearize tabular data into text with randomly permuted feature orders, and then feed the sequence into a language model. A dynamic token-level dependency graph is learned during training, guided by soft constraints from functional dependencies, and used to modulate attention weights for structure-aware tabular modeling.}
    \label{fig:pipeline}
\end{figure*}

\paragraph{Attention as Graph Structure}
In transformer-based language models, the self-attention mechanism can be treated as a fully connected graph over input tokens \citep{yao2020multimodal}. For a sequence of tokens $\{t_1, t_2, ..., t_n\}$, the attention weight $a_{ij}$ between tokens $t_i$ and $t_j$ is computed as:
\begin{equation}
a_{\text{ij}} = \text{softmax}\left(\frac{\mathbf{q}_i\mathbf{k}_j^\top}{\sqrt{d_k}}\right),
\end{equation}
where $\mathbf{q}_i$ and $\mathbf{k}_j$ are the query and key vectors associated with tokens $t_i$ and $t_j$. This mechanism can be interpreted as defining a fully connected directed graph $G_{\text{LM}} = (V, E)$, where $V = \{t_1, ..., t_n\}$ represents the tokens, and $E = \{(t_i, t_j) \mid i, j \in [1, n] \}$ denotes edges with weights $a_{ij}$. This structure assumes that any token can directly influence any other token, which is suitable for natural language where long-range dependencies are frequent and contextually relevant \citep{chaudhari2021attentive}.

\paragraph{Relational Structure of Tabular Data}
In contrast, tabular data typically exhibits sparse relational structures, where each feature depends only on a small subset of other features \citep{papenbrock2015functional}. Such relationships can be represented as a sparse graph $G_{\text{Tab}} = (V', E')$, where vertices $V' = \{f_1, f_2, ..., f_d\}$ represent features, and edges $E' \subset V' \times V'$ represent pairwise dependencies, with $|E'| \ll |V'|^2$. Formally, the generative process of each feature value $v_i$ can be described as:
\begin{equation}
    v_i = \psi_i\left(N(i), \varepsilon_i\right),
\end{equation}
where $N(i)$ is a small set of features that $f_i$ depends on, $\psi_i$ is a functional mapping, and $\varepsilon_i$ denotes independent noise. As empirically observed by \citet{liu2023goggle}, such sparse relational structures are characteristic of real-world data generation processes.

\paragraph{Structure Mismatch Problem}
When applying language models to tabular data generation, a critical mismatch arises between the dense attention structure $G_{\text{LM}}$ and the sparse relational structure $G_{\text{Tab}}$. This mismatch leads to \textbf{two key problems}:  
(1) The attention mechanism assigns non-negligible weights across all token pairs, diluting its ability to capture the few critical dependencies as the number of features increases;  
(2) Language models do not encode any inductive bias about tabular feature dependencies, and must learn such structural patterns purely from data, which becomes increasingly difficult as table complexity grows.

Motivated by these limitations, we propose to explicitly incorporate functional dependency structures into language models, as detailed in the following sections.

\section{Graph-Guided Dependency Learning}
\label{sec: Methodology}
We propose GraDe, a novel method that enhances language models to better capture the sparse dependency structures inherent in tabular data. Building on the observation that tabular data contains natural sparsity in feature relationships, GraDe integrates this structural information into the attention mechanism of language models.

\subsection{Overview}
Recent work has established sparse dependencies in tabular data can be more accurately captured through relational structures \citep{liu2023goggle}. While language models' self-attention mechanisms create relational representations among tokens, we hypothesize that their dense connectivity pattern, where every token attends to all others, may dilute focus on the critical relationships in tabular data.

GraDe addresses this challenge by learning a dynamic sparse graph that guides the attention mechanism to prioritize structurally important dependencies. Our approach incorporates three key innovations: (1) a graph-guided attention that dynamically models sparse token-level relationships; (2) integration of externally derived functional dependencies as supervision; and (3) sparsity regularization to focus attention on meaningful connections. Unlike previous methods, GraDe introduces minimal modifications to only the attention modules while preserving the semantic capabilities of pretrained language models. Figure \ref{fig:pipeline} shows our architecture.

\subsection{Dynamic Graph-Guided Attention}
\label{subsec:gynamic_graph_attn}
\paragraph{Graph Construction}
For a sequence of tokens, GraDe learns a weighted directed graph for each attention head, represented by an adjacency matrix $\mathbf{W} \in [0,1]^{n \times n}$, where each entry $w_{ij}$ quantifies the importance of the directed relation from token $j$ to token $i$. The graph is parameterized by a lightweight two-layer neural network $G_\phi$:
\begin{align}
w_{ij} &= G_\phi(q_i, k_j) \\
       &= \sigma\left(W_2 \cdot \text{ReLU}(W_1 \cdot [q_i; k_j])\right),
\end{align}
where $q_i, k_j \in \mathbb{R}^{d_k}$ are the query and key vectors, $W_1 \in \mathbb{R}^{d_k \times 2d_k}$ and $W_2 \in \mathbb{R}^{1 \times d_k}$ are learnable parameters, and $\sigma$ denotes the sigmoid activation. This module operates over the head dimension $d_k$, which leads to parameter efficiency.

\paragraph{Attention Modulation}
Standard self-attention computes alignment scores via query-key dot products \citep{brauwers2021general}, treating all token pairs as equally relevant candidates for attention. To encourage the model to focus on important relations, GraDe modulates these scores by incorporating the learned dependency weights through a logarithmic gating mechanism:
\begin{equation}
\tilde{a}_{ij} = a_{ij} \cdot \log(w_{ij} + \epsilon),
\end{equation}
where $a_{ij}$ is the unnormalized attention score, $w_{ij}$ is the learned edge weight, and $\epsilon$ is a small constant for numerical stability. This transformation acts as a gating mechanism: when $w_{ij} \approx 1$ (indicating a strong dependency), it preserves the original score; when $w_{ij} \approx 0$, it strongly suppresses it. The modulated attention scores $\tilde{a}_{ij}$ are subsequently normalized via softmax, ensuring a valid attention distribution.

During inference, we optimize graph computation for the autoregressive setting by computing only the edges involving the newly generated token, enabling efficient decoding for long sequences.

\subsection{Incorporating Functional Dependencies}
\label{subsec:incorporating_fd}
While the dynamic graph module captures token-level dependencies within the sequence, tabular data primarily exhibits important relationships at the feature level \citep{yan2024making}. These relationships often take the form of functional dependencies, where one set of features uniquely determines another.

To bridge this gap between token-level and feature-level dependencies, we incorporate prior knowledge about FDs into the training objective. For each known FD $X \rightarrow Y$, we compute the average connection strength from tokens representing features in $X$ to those representing features in $Y$:
\begin{equation}
\bar{w}_{Y,X} = \frac{1}{|T_Y||T_X|} \sum_{i \in T_Y} \sum_{j \in T_X} w_{ij},
\end{equation}
where $T_X$ and $T_Y$ are the sets of token indices corresponding to features $X$ and $Y$, respectively.

Based on these average connection strengths, we define an FD alignment loss:
\begin{equation}
\mathcal{L}_{\text{FD}} = \sum_{X \rightarrow Y \in \mathcal{F}} \phi(\bar{w}_{Y,X}, \alpha),
\label{equ:loss_FD}
\end{equation}
where $\mathcal{F}$ is the set of known FDs and $\phi$ is a constraint function enforcing a minimum strength threshold $\alpha$.

We implement this constraint using a smooth softplus formulation:
\begin{equation}
\phi(w, \alpha) = \log(1 + \exp(\alpha - w)),
\end{equation}
This formulation enables flexible modeling of functional dependencies while accommodating the inherent variability in real-world data. By smoothly penalizing insufficient connection strengths, the model is encouraged to focus on key structural relationships without imposing overly rigid constraints.

\subsection{Multi-Objective Fine-Tuning}
\label{subsec:fine_tuning}
GraDe is trained using a composite loss that jointly optimizes for language modeling quality, structural sparsity, and alignment with known dependencies:

\paragraph{Language Modeling}
To preserve the sequential generation capability of language models, we adopt standard autoregressive training with the cross-entropy loss:
\begin{equation}
\mathcal{L}_{\text{LM}} = -\sum_{i=1}^{|x|} \log P(x_i \mid x_{<i}),
\end{equation}
where $x$ is the linearized input sequence derived from tabular data.

\paragraph{Graph Sparsity Regularization}
To reflect the inherent sparsity in tabular data relationships, we impose an L1 penalty on the learned adjacency matrices:
\begin{equation}
\mathcal{L}_{\text{sparse}} = \frac{1}{L} \sum_{l=1}^{L} \| \mathbf{W}^{(l)} \|_1,
\end{equation}
where $L$ is the number of transformer layers and $\mathbf{W}^{(l)}$ is the graph matrix at layer $l$. This regularization encourages the model to assign non-zero weights only to structurally meaningful dependencies, and avoid dense attention allocation.

\paragraph{FD Alignment}
The FD alignment loss (as given in Eq.~\ref{equ:loss_FD}) guides the model to respect known feature-level constraints by encouraging strong connections between dependent feature tokens.

\paragraph{Overall Training Loss}
The overall training loss combines the language modeling objective, sparsity regularization, and FD alignment:
\begin{equation}
\mathcal{L} = \mathcal{L}_{\text{LM}} + \lambda_{\text{sparse}} \cdot \mathcal{L}_{\text{sparse}} + \lambda_{\text{FD}} \cdot \mathcal{L}_{\text{FD}},
\label{equ:overall_loss}
\end{equation}
where $\lambda_{\text{sparse}}$ and $\lambda_{\text{FD}}$ are hyperparameters that control the influence of sparsity and dependency alignment, respectively. 

This multi-objective approach enables GraDe to learn representations that are both effective for language modeling and structurally aligned with the underlying tabular data dependencies.

\begin{table*}[ht] 
    \centering
    \renewcommand{\arraystretch}{1.05}
    \footnotesize
    {
        \resizebox{\textwidth}{!}{
	\begin{tabular}{rlc|ccccc|cc}
            \toprule[0.8pt]
            Dataset & & Original & TVAE & CTGAN & CTABGAN+ & TabSyn & GReaT & \textbf{GraDe} & \textbf{GraDe-Light}\\
            \midrule 
            \multirow{3}{*}{Bird ($\uparrow$)} & DT & $99.97${\tiny$\pm0.02$}  & $70.59${\tiny$\pm0.00$} & $94.28${\tiny$\pm0.17$} & $66.46${\tiny$\pm0.21$} & $99.75${\tiny$\pm0.06$}  & $99.63${\tiny$\pm0.07$}  & $99.70${\tiny$\pm0.00$}  & \redbf{$99.78$}{\tiny\redbf{$\pm0.02$}} \\ 
            & RF & $100.00${\tiny$\pm0.00$} & $79.87${\tiny$\pm0.46$} & $99.09${\tiny$\pm0.08$}  & $75.54${\tiny$\pm0.11$} & $99.97${\tiny$\pm0.12$}  & \redbf{$100.00$}{\tiny\redbf{$\pm0.00$}} & \redbf{$100.00$}{\tiny\redbf{$\pm0.00$}} & \redbf{$100.00$}{\tiny\redbf{$\pm0.00$}} \\
            & LR & $100.00${\tiny$\pm0.00$} & $87.04${\tiny$\pm0.00$} & \redbf{$100.00$}{\tiny\redbf{$\pm0.00$}} & $73.99${\tiny$\pm0.00$} & \redbf{$100.00$}{\tiny\redbf{$\pm0.00$}} & $98.29${\tiny$\pm0.00$}  & \redbf{$100.00$}{\tiny\redbf{$\pm0.00$}} & $98.85${\tiny$\pm0.00$} \\
            \midrule
            
            \multirow{3}{*}{Sick ($\uparrow$)} & DT & $98.70${\tiny$\pm0.10$} & $95.87${\tiny$\pm0.26$} & $86.28${\tiny$\pm1.20$} & $92.40${\tiny$\pm0.42$} & $74.65${\tiny$\pm1.05$} & $91.42${\tiny$\pm0.61$} & \redbf{$96.05$}{\tiny\redbf{$\pm0.42$}} & $92.19${\tiny$\pm0.37$} \\
            & RF & $98.46${\tiny$\pm0.18$} & $95.34${\tiny$\pm0.24$} & $94.97${\tiny$\pm0.00$} & $94.99${\tiny$\pm0.26$} & $96.21${\tiny$\pm0.34$} & $96.40${\tiny$\pm0.15$} & \redbf{$98.04$}{\tiny\redbf{$\pm0.15$}} & $97.01${\tiny$\pm0.22$}\\
            & LR & $89.54${\tiny$\pm0.00$} & \redbf{$94.57$}{\tiny\redbf{$\pm0.00$}} & $58.01${\tiny$\pm0.00$} & $82.65${\tiny$\pm0.00$} & $66.23${\tiny$\pm0.00$} & $83.31${\tiny$\pm0.00$} & $90.60${\tiny$\pm0.00$} & $91.72${\tiny$\pm0.00$}\\
            \midrule   
            
            \multirow{3}{*}{Income ($\uparrow$)} & DT & $81.14${\tiny$\pm0.03$} & $80.28${\tiny$\pm0.12$} & $79.87${\tiny$\pm0.17$} & $76.61${\tiny$\pm0.31$} & $80.73${\tiny$\pm0.13$} & $79.09${\tiny$\pm0.14$} & \redbf{$80.94$}{\tiny\redbf{$\pm0.19$}} & $79.63${\tiny$\pm0.12$} \\
            & RF & $85.15${\tiny$\pm0.15$} & $82.62${\tiny$\pm0.13$} & $82.25${\tiny$\pm0.13$} & $83.38${\tiny$\pm0.18$} & $80.22${\tiny$\pm0.11$} & $83.68${\tiny$\pm0.12$} & \redbf{$84.05$}{\tiny\redbf{$\pm0.06$}} & $83.65${\tiny$\pm0.19$}\\
            & LR & $80.45${\tiny$\pm0.00$} & $78.99${\tiny$\pm0.00$} & $79.19${\tiny$\pm0.00$} & $77.07${\tiny$\pm0.00$} & \redbf{$81.14$}{\tiny\redbf{$\pm0.00$}} & $80.19${\tiny$\pm0.00$} & $80.94${\tiny$\pm0.00$} & $80.02${\tiny$\pm0.00$}\\
            \midrule
         
            \multirow{3}{*}{Diabetes ($\uparrow$)} & DT & $74.68${\tiny$\pm1.30$} & $72.21${\tiny$\pm0.49$} & $59.48${\tiny$\pm0.88$} & $62.99${\tiny$\pm1.23$} & $75.06${\tiny$\pm2.08$} & $65.97${\tiny$\pm0.66$} & \redbf{$78.05$}{\tiny\redbf{$\pm0.66$}} & $67.08${\tiny$\pm1.40$} \\
            & RF & $74.94${\tiny$\pm0.88$} & $75.97${\tiny$\pm0.71$} & $57.01${\tiny$\pm1.50$} & $64.81${\tiny$\pm1.50$} & $75.32${\tiny$\pm0.92$} & $67.40${\tiny$\pm0.76$} & \redbf{$77.27$}{\tiny\redbf{$\pm1.42$}} & $71.56${\tiny$\pm1.76$} \\
            & LR & $69.48${\tiny$\pm0.00$} & \redbf{$73.38$}{\tiny\redbf{$\pm0.00$}} & $57.14${\tiny$\pm0.00$} & $72.08${\tiny$\pm0.00$} & $70.78${\tiny$\pm0.00$} & $67.53${\tiny$\pm0.00$} & $68.83${\tiny$\pm0.00$} & $66.88${\tiny$\pm0.00$} \\
            \midrule

            \multirow{3}{*}{Housing ($\downarrow$)} & DT & $0.24${\tiny$\pm0.01$} & $0.35${\tiny$\pm0.00$} & $0.50${\tiny$\pm0.00$} & $0.39${\tiny$\pm0.00$} & $0.30${\tiny$\pm0.01$} & $0.34${\tiny$\pm0.00$} & \redbf{$0.27$}{\tiny\redbf{$\pm0.00$}} & $0.31${\tiny$\pm0.01$} \\
            & RF & $0.18${\tiny$\pm0.00$} & $0.30${\tiny$\pm0.01$} & $0.37${\tiny$\pm0.00$} & $0.28${\tiny$\pm0.01$} & $0.22${\tiny$\pm0.00$} & $0.24${\tiny$\pm0.01$} & \redbf{$0.21$}{\tiny\redbf{$\pm0.01$}}  & $0.23${\tiny$\pm0.00$} \\
            & LR & $0.29${\tiny$\pm0.00$} & $0.33${\tiny$\pm0.00$} & $0.35${\tiny$\pm0.00$} & $0.29${\tiny$\pm0.00$} & \redbf{$0.29$}{\tiny\redbf{$\pm0.00$}} & $0.30${\tiny$\pm0.00$} & $0.31${\tiny$\pm0.00$} & $0.31${\tiny$\pm0.00$} \\
		\bottomrule[1.0pt] 
		\end{tabular}}
  }
  \caption{Machine learning efficiency experiment. The best results are marked in \redbf{Red}. ``Original'' refers to training on original real data. Classification accuracy ($\uparrow$) and regression mean absolute percentage error ($\downarrow$) are reported.} 
  \label{tab:mle}
\end{table*}

\paragraph{Parameter-Efficient Variant}
We also introduce GraDe-Light, a lightweight variant that updates only the dynamic graph-guided attention modules while freezing all other parameters. This approach significantly reduces trainable parameters while maintaining most of the model's dependency modeling capability. We evaluate both variants to assess trade-offs between computational efficiency and generation quality.


\section{Experimental Setup}
\label{sec:experimental_setup}
\paragraph{Models} 
We use GPT-2 \citep{radford2019language} with 124 million parameters as our backbone model for all main experiments. We evaluate both the full GraDe and GraDe-Light variants on this model. Additional experiments with GPT-2 Medium with 355 million parameters are reported in Appendix~\ref{subsec:mle_gpt2_m}. We use nucleus sampling \citep{holtzman2019curious} for text generation.

\paragraph{Datasets}
For our main experiments, we use five real-world datasets from diverse domains with complex dependencies: \textbf{Bird} (Ecology), \textbf{Sick} and \textbf{Diabetes} (Medical), \textbf{Income} (Social), and \textbf{Housing} (Real Estate). Dataset sizes range from 700 to 40,000 samples with 5 to 30 features, including continuous, categorical, and integer types. For our scaling experiments (\S\ref{subsec:scaling}), we use the \textbf{Loan} dataset, a large-scale financial dataset with approximately 250,000 samples. Detailed descriptions and the dataset statistics are provided in Appendix~\ref{subsec:datasets}.

\paragraph{FDs Extraction}
Functional dependency detection is a mature, well-studied problem in the database domain \citep{liu2010discover}. We use HyFD \citep{papenbrock2016hybrid}, a hybrid algorithm for functional dependency discovery, to automatically extract FDs from our datasets. We verify the results using TANE \citep{huhtala1999tane}. The detected FDs serve as prior knowledge for tabular data modeling. Further details on the detection algorithms and the extracted FDs for each dataset are provided in Appendix~\ref{subsec:fds}.

\paragraph{Baselines}
We compare GraDe with five representative tabular data generation methods: VAE-based TVAE \citep{xu2019modeling}, which employs mode-specific normalization; GAN-based methods CTGAN \citep{xu2019modeling} and CTABGAN+ \citep{zhao2024ctab}, where CTGAN uses mode-specific normalization for handling complex distributions and CTABGAN+ introduces downstream losses and specialized encoders for mixed variables; diffusion-based method TabSyn \citep{zhang2023mixed}, which operates in a latent space to capture inter-column relationships; and LLM-based method GReaT \citep{borisov2022language}.

\paragraph{Evaluation}
Following prior works on tabular data generation \citep{borisov2022language, gulati2023tabmt, liu2023goggle}, we evaluate synthetic data quality across three key dimensions: (1) \textbf{Utility}: we assess whether synthetic data can effectively replace real data by training models on synthetic data and evaluating them on real test data \citep{esteban2017real}; (2) \textbf{Fidelity}: we measure how well the statistical dependencies between attributes are preserved, which is often more critical than matching individual distributions; and (3) \textbf{Privacy}: we verify that generated data resembles but doesn't duplicate training examples by computing distances between synthetic samples and their closest records in the original dataset. For all evaluations, we report average results over five runs with different random seeds to ensure reproducibility.


\section{Results}
\label{sec:results}
\subsection{Synthetic Data Quality}

\begin{figure*}
    \centering
    \includegraphics[width=1\linewidth]{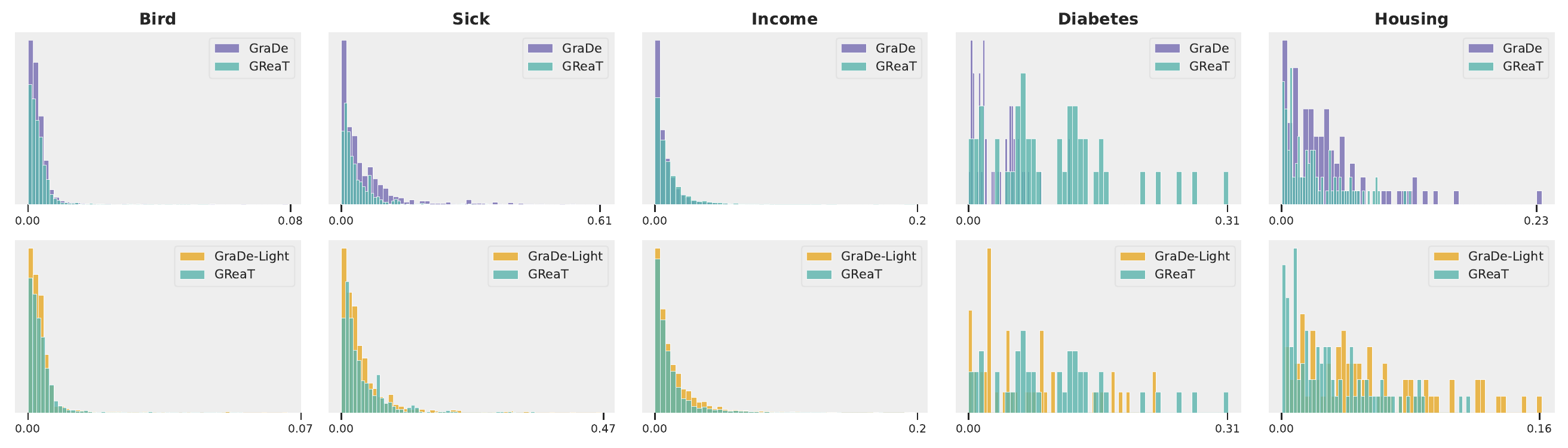}
    \caption{Correlation error histograms across five datasets. Each bar represents the absolute difference between real and synthetic data correlation coefficients, with values closer to zero indicating better fidelity. GraDe (\textcolor{purple}{purple}) and GraDe-Light (\textcolor{yellow}{yellow}) show error distributions more concentrated near zero compared to GReaT (\textcolor{green}{green}).}
    \label{fig:correlation_error}
\end{figure*}

\begin{table*}[ht] 
    \centering
    \renewcommand{\arraystretch}{1.1}
    \footnotesize
    {
	\begin{tabular}{lccccc|c}
            \toprule[0.8pt]
            Method & Bird & Sick & Income & Diabetes & Housing & Average \\
            \midrule 
            TVAE & $.434${\tiny$\pm.032$} & $.262${\tiny$\pm.023$} & $.156${\tiny$\pm.025$}  & $.360${\tiny$\pm.010$} & $.165${\tiny$\pm.006$} & $.275${\tiny$\pm.019$} \\
            CTGAN & $.180${\tiny$\pm.022$} & $.702${\tiny$\pm.040$} & $.560${\tiny$\pm.044$}  & $.783${\tiny$\pm.020$} & $.259${\tiny$\pm.010$} & $.497${\tiny$\pm.027$} \\
            CTABGAN+ & $.279${\tiny$\pm.027$} & $.386${\tiny$\pm.032$} & $.533${\tiny$\pm.036$}  & $.516${\tiny$\pm.014$} & $.197${\tiny$\pm.008$} & $.382${\tiny$\pm.023$} \\
            TabSyn & $.008${\tiny$\pm.004$} & $.202${\tiny$\pm.019$} & $.397${\tiny$\pm.033$} & $.385${\tiny$\pm.014$} & $.152${\tiny$\pm.006$} & $.229${\tiny$\pm.015$} \\
            GReaT & $.010${\tiny$\pm.005$} & $.098${\tiny$\pm.011$} & $.165${\tiny$\pm.059$}  & $.368${\tiny$\pm.012$} & $.117${\tiny$\pm.004$} & $.152${\tiny$\pm.018$} \\
            \midrule
            \textbf{GraDe} & \redbf{$.002$}{\tiny\redbf{$\pm.001$}} & \redbf{$.081$}{\tiny\redbf{$\pm.009$}} & $.161${\tiny$\pm.019$}  & \redbf{$.323$}{\tiny\redbf{$\pm.012$}} & \redbf{$.103$}{\tiny\redbf{$\pm.003$}} & \redbf{$.134$}{\tiny\redbf{$\pm.009$}} \\
            \textbf{GraDe-Light} & $.007${\tiny$\pm.004$} & $.145${\tiny$\pm.065$} & \redbf{$.147$}{\tiny\redbf{$\pm.018$}}  & $.355${\tiny$\pm.014$} & $.112${\tiny$\pm.003$} & $.153${\tiny$\pm.021$} \\
		\bottomrule[1.0pt] 
		\end{tabular}
  }
  \caption{Distance to closest record (DCR) scores (lower is better). The best results are marked in \redbf{Red}. } 
  \label{tab:dcr}
  \vspace{-3pt}
\end{table*}

\paragraph{Machine Learning Efficiency (MLE)} 
We evaluate the practical \textit{utility} of synthetic data by training discriminative models on the generated datasets and testing their performance on real test set. This setup reflects whether the synthetic data can effectively replace real data in downstream learning tasks. To ensure generality across model classes, we use three commonly adopted classifiers: Decision Tree (DT), Random Forest (RF)~\citep{ho1995random}, and Linear/Logistic Regression (LR). Table \ref{tab:mle} shows MLE scores across datasets compared to models trained on the original training dataset. GraDe demonstrates strong performance, with particularly significant improvements over GReaT (our closest LLM-based competitor) on complex medical datasets, achieving gains of up to 12\% on Diabetes and 5\% on Sick. These results highlight the advantage of explicit dependency modeling in domains with complex structural relationships.  Our GraDe-Light variant achieves similar performance while using 100 million fewer trainable parameters than GReaT, highlighting the efficiency of our graph-guided approach.

\paragraph{Column-Wise Correlation Error}
The preservation of feature relationships is crucial for generating realistic tabular data. Following \citet{gulati2023tabmt}, we assess \textit{fidelity} by measuring the absolute difference between Pearson correlation coefficients in real versus synthetic data. For categorical features, we first apply one-hot encoding before calculating correlations. Figure \ref{fig:correlation_error} shows the correlation error histograms comparing GraDe, GraDe-Light, and GReaT across different datasets. We focus this comparison on LLM-based approaches since prior work has shown that autoregressive models typically struggle with capturing joint distributions \citep{zhang2023mixed}, making this a particularly challenging aspect for LLM-based tabular data generation. Our results demonstrate that GraDe's graph-guided approach provides a clear advantage by explicitly modeling functional dependencies and directing attention to structurally relevant connections.

\paragraph{Distance to the Closest Record (DCR)}
\textit{Privacy} preservation is a key application for synthetic tabular data, requiring generated samples to resemble but not duplicate training examples. Following \citet{zhang2023mixed}, we use the L1 norm to measure the minimum distance between each synthetic sample and real training examples. For numerical features, differences are normalized by the observed feature range; for categorical features, the distance is defined as 0 if the values match, and 1 otherwise. The final DCR score is computed as the average of the minimum distances across all synthetic samples. Table~\ref{tab:dcr} reports the DCR scores across all five datasets. While most models perform adequately on simpler datasets, our approach shows substantial improvements on medical datasets (Sick, Diabetes) with complex dependencies. This enhancement is particularly valuable in sensitive domains, as lower DCR scores indicate greater dissimilarity from training records and are associated with reduced membership inference risks~\citep{hu2022membership}. These results demonstrate how explicit dependency modeling enables language models to better balance fidelity and privacy.

\subsection{Modeling in Low-Data Regimes}
\label{subsec:low_data}
\begin{figure}[htbp]
    \centering
    \includegraphics[width=0.95\linewidth]{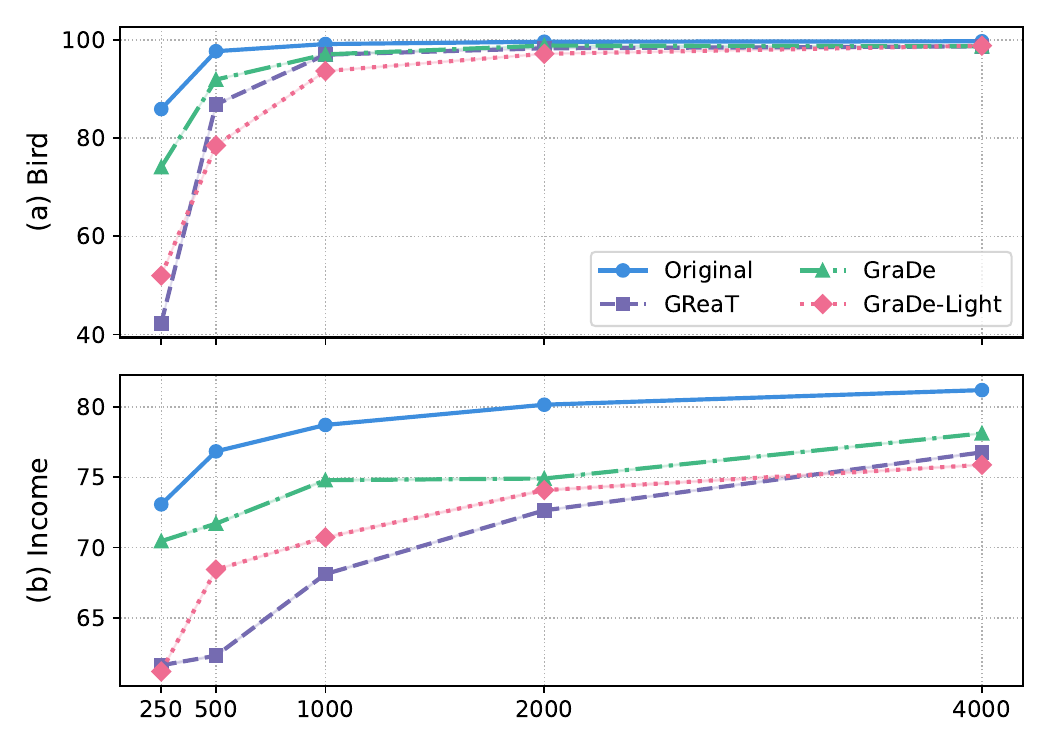}
    \caption{Classification accuracy across different training sample sizes in low-data regimes. ``Original'' denotes models trained on original real data; others are trained on synthetic data.}
    \label{fig:low_data_acc}
    \vspace{-12pt}
\end{figure}

Data-oriented generative approaches like GAN or diffusion-based methods typically struggle in data-scarce scenarios, requiring substantial training examples to capture complex distributions \citep{manousakas2023usefulness}. In contrast, LLM-based methods can leverage rich pre-trained knowledge to model tabular data effectively even with limited samples \citep{seedat2023curated, gardner2024large}. We evaluate this capability on Bird and Income datasets using training sets from 250 to 4000 examples, comparing GraDe against GReaT on identical data subsets. Figure~\ref{fig:low_data_acc} shows GraDe consistently outperforms GReaT across all data regimes, with gains up to 15\% under extreme scarcity (250 examples). Although the gap narrows with more data, GraDe maintains its advantage throughout. GraDe-Light achieves comparable improvements despite fewer parameters. We observe more pronounced improvements on the Income dataset, where the extracted functional dependencies are both more numerous and structurally complex. This empirical pattern suggests that GraDe benefits more from structural guidance when the underlying dependencies are richer or harder to learn implicitly.

\subsection{Ablation Study}
\label{subsec:ablation}
We assess the contribution of each component in our approach through ablation experiments with two variants: (1) GraDe w/o Sparsity, which removes graph sparsity regularization, and (2) GraDe w/o FDs, which removes FD alignment loss. Table~\ref{tab:ablation} compares these variants against the full GraDe model and GReaT across all datasets using decision trees trained on generated data.

\begin{table}[htbp] 
    \centering
    \renewcommand{\arraystretch}{1.1}
    \footnotesize
    {
	\resizebox{\linewidth}{!}{
        \begin{tabular}{rccccc}
            \toprule[1.0pt]
             & \textbf{Bird} & \textbf{Sick} & \textbf{Income} & \textbf{Diabetes} & \textbf{Housing} \\
             & Acc $\uparrow$ & Acc $\uparrow$ & Acc $\uparrow$ & Acc $\uparrow$ & MAPE $\downarrow$ \\ 
            \midrule[0.3pt] 
            GraDe & \:$99.70$ & \:$96.05$ & \:$80.94$ & \:$78.05$ & \:$0.27$ \\
            \midrule[0.3pt]  
             w/o Sparsity & $+0.23$ & $-0.40$ & $-0.51$ & $-0.63$ & $+0.01$ \\
             w/o FDs & $-0.03$ & $-6.51$ & $-0.51$ & $-5.47$ & $-0.03$  \\
            GReaT & $-0.07$ & $-4.63$ & $-1.85$ & $-12.08$ & $-0.07$ \\
		\bottomrule[1.0pt] 
		\end{tabular}}
  }
  \caption{Ablation study across five datasets comparing GraDe, its ablated variants, and the GReaT baseline. Accuracy (Acc) is used for classification tasks and mean absolute percentage error (MAPE) for regression tasks.} 
  \label{tab:ablation}
  \vspace{-10pt}
\end{table}

Both components enhance GraDe's effectiveness, with FD alignment having the stronger impact. Removing FD alignment significantly degrades performance on medical datasets (Sick and Diabetes), where numerical attributes and ambiguous feature names make implicit learning challenging. For datasets with clearer dependencies (Bird, Income, Housing), the impact is smaller, and removing sparsity regularization occasionally improves performance, possibly because the attention distribution benefits from added flexibility in datasets with strong local patterns. These results demonstrate that explicit dependency guidance is most valuable for complex data structures, while both components work complementarily to optimize overall performance, providing empirical support for our hypothesis that dense attention may dilute focus on critical relationships.

\section{Conclusion}
In this work, we address the structural mismatch between LLMs' dense attention patterns and tabular data's sparse dependencies. Our solution, GraDe, guides token interactions through learnable dependency graphs that integrate functional relationships into attention mechanisms. This approach helps models focus on meaningful connections without altering the base architecture. Experiments across diverse datasets show consistent gains in utility, fidelity, and privacy. GraDe outperforms prior LLM-based methods by up to 12\% on complex medical datasets and remains effective even with only 250 training samples. The lightweight GraDe-Light variant achieves similar results while reducing trainable parameters by nearly 100 million. These findings confirm that structural inductive bias significantly enhances LLMs for tabular modeling. By bridging the gap between powerful semantic representations and domain-specific structural awareness, GraDe offers a practical and effective solution for generating high-quality synthetic tabular data.

\section{Limitations}
While GraDe effectively enhances tabular data modeling through functional dependencies, several limitations exist. First, extremely high-dimensional tables may exceed LLM context limits and increase computational costs. While GraDe-Light reduces parameters, sequential generation of wide tables remains inefficient. Future models with larger contexts may naturally address these scaling issues.

Second, our method relies on externally extracted functional dependencies, which may inherit dataset biases or miss subtle relationships. However, automatic FD extraction provides scalable, domain-agnostic supervision without expert annotation. GraDe mitigates bias through soft constraints, though critical applications may benefit from expert verification.

Finally, training converges slower than baseline models because our graph-guided attention modules train from scratch while language model components use pretrained weights. The graph structures need additional iterations to align with pretrained representations. Future work could explore transfer learning or better initialization for faster convergence.

\section*{Acknowledgments}
This work was partially supported by the Verband der Vereine Creditreform e.V.. We thank Andreas Kotowski and his colleagues from Creditreform for their valuable collaboration and for providing useful discussions and insights that contributed to this work.

\paragraph{Use of AI Assistants} The authors acknowledge the use of ChatGPT exclusively to refine the text in the final manuscript.

\bibliography{custom}

\newpage
\appendix

\section{Additional Results}
\label{sec:additional_results}
This section extends our evaluation with supplementary experiments, providing deeper insights into GraDe's effectiveness across diverse scenarios. Specifically, we evaluate the approach on a large-scale imbalanced dataset (\S\ref{subsec:scaling}), measure intrinsic constraint preservation in generated data (\S\ref{subsec:constraint_fidelity}), and investigate scalability to a larger backbone model (\S\ref{subsec:mle_gpt2_m}). Additionally, we assess data fidelity using discriminator-based measures (\S\ref{subsec:discriminator}), examine sensitivity to functional dependency alignment weights (\S\ref{subsec:impact_fd}), and evaluate the impact of sparsity constraint on high-dimensional data (\S\ref{subsec:ablation_wide}).

\subsection{Scaling to Large-Scale Imbalanced Data}
\label{subsec:scaling}
Real-world large-scale datasets, particularly in financial domains such as loan default prediction, commonly exhibit severe imbalance issues, with minority classes often being critically important yet underrepresented. To evaluate GraDe's performance on such challenges, we conducted additional machine learning efficiency experiments on a large-scale loan dataset containing approximately 250,000 samples with only 12\% positive cases predicting loan defaults based on customer behavior. This experiment directly assesses the utility of synthetic data generated by different methods when applied to large imbalanced datasets.

Table~\ref{tab:mle_loan} presents AUC and F1 scores as evaluation metrics, which better represent performance on imbalanced data compared to accuracy. The results reveal distinct performance patterns across different methods. TVAE completely failed to generate minority class samples, likely due to posterior collapse \citep{lucas2019understanding}. While CTGAN and CTABGAN+ managed to capture some minority patterns, their performance remains limited with AUC scores barely above random chance. TabSyn demonstrates strong performance, particularly with RF and LR models, leveraging its latent space modeling approach. GraDe achieves the best performance with DT models and remains competitive across other classifiers, outperforming other LLM-based methods. These results highlight GraDe's effectiveness at capturing complex dependencies even in challenging imbalanced scenarios, where modeling minority classes accurately is crucial for downstream applications.

\begin{table*}[ht] 
    \centering
    \renewcommand{\arraystretch}{1.02}
    \footnotesize
    {
	\begin{tabular}{lcccccc}
        \toprule
        & \multicolumn{2}{c}{\textbf{DT}} & \multicolumn{2}{c}{\textbf{RF}} & \multicolumn{2}{c}{\textbf{LR}} \\
        \cmidrule(lr){2-3} \cmidrule(lr){4-5} \cmidrule(lr){6-7}
                        & AUC $\uparrow$  & F1 $\uparrow$  & AUC $\uparrow$  & F1 $\uparrow$  & AUC $\uparrow$  & F1 $\uparrow$  \\
        \midrule
        Original    & $.739${\tiny$\pm.000$} & $.347${\tiny$\pm.000$} & $.814${\tiny$\pm.002$} & $.434${\tiny$\pm.003$} & $.635${\tiny$\pm.000$} & $.268${\tiny$\pm.000$} \\
        \midrule
        TVAE        & \multicolumn{1}{c}{-}  & \multicolumn{1}{c}{-}  & \multicolumn{1}{c}{-}  & \multicolumn{1}{c}{-}  & \multicolumn{1}{c}{-}  & \multicolumn{1}{c}{-}  \\
        CTGAN       & $.505${\tiny$\pm.001$} & $.211${\tiny$\pm.001$} & $.522${\tiny$\pm.001$} & $.210${\tiny$\pm.002$} & $.521${\tiny$\pm.000$} & $.210${\tiny$\pm.000$} \\
        CTABGAN+    & $.521${\tiny$\pm.001$} & $.217${\tiny$\pm.000$} & $.538${\tiny$\pm.002$} & $.222${\tiny$\pm.001$} & $.543${\tiny$\pm.000$} & $.224${\tiny$\pm.000$} \\
        TabSyn      & \underline{$.571${\tiny$\pm.001$}} & $.246${\tiny$\pm.001$} & \redbf{$.645$}{\tiny\redbf{$\pm.001$}} & \redbf{$.269$}{\tiny\redbf{$\pm.002$}} & \redbf{$.594$}{\tiny\redbf{$\pm.000$}} & \redbf{$.244$}{\tiny\redbf{$\pm.000$}} \\
        GReaT       & $.569${\tiny$\pm.000$} & $.254${\tiny$\pm.000$} & $.622${\tiny$\pm.002$} & $.252${\tiny$\pm.002$} & $.553${\tiny$\pm.000$} & $.224${\tiny$\pm.000$} \\
        \midrule
        \textbf{GraDe} & \redbf{$.587$}{\tiny\redbf{$\pm.001$}} & \redbf{$.271$}{\tiny\redbf{$\pm.000$}} & \underline{$.642${\tiny$\pm.003$}} & \underline{$.267${\tiny$\pm.004$}} & \underline{$.581${\tiny$\pm.000$}} & \underline{$.238${\tiny$\pm.000$}} \\
        \textbf{GraDe-Light} & $.576${\tiny$\pm.001$} & \underline{$.260${\tiny$\pm.000$}} & $.611${\tiny$\pm.004$} & $.245${\tiny$\pm.002$} & $.550${\tiny$\pm.000$} & $.218${\tiny$\pm.000$} \\
        \bottomrule
        \end{tabular}
  }
  \caption{Additional results for machine learning experiment on the large-scale imbalanced \textbf{Loan} dataset (with approx. 250K samples and 12\% positive class). The best results are marked in \redbf{Red}, second-best results are \underline{underlined}. We report AUC and F1 scores here ($\uparrow$).} 
  \vspace{-5pt}
  \label{tab:mle_loan}
\end{table*}

\begin{table*}[ht] 
    \centering
    \renewcommand{\arraystretch}{1.05}
    \footnotesize
    {
	\begin{tabular}{lccc|c}
            \toprule[0.8pt]
            \multirow{2}{*}{Method} & Bird & Income & Housing & \multirow{2}{*}{Average} \\
            & \scriptsize lat, long $\rightarrow$ State & \scriptsize education $\rightarrow$ education\_num & \scriptsize latitude, longitude $\rightarrow$ CA & \\
            \midrule 
            TVAE & $15.75${\tiny$\pm0.63$} & $18.76${\tiny$\pm0.47$} & $14.14${\tiny$\pm0.53$} & $16.22${\tiny$\pm0.54$} \\
            CTGAN & $21.09${\tiny$\pm0.70$} & $20.64${\tiny$\pm0.49$} & $32.91${\tiny$\pm0.72$} & $24.88${\tiny$\pm0.64$} \\
            CTABGAN+ & $16.78${\tiny$\pm0.65$} & $14.35${\tiny$\pm0.43$} & $17.15${\tiny$\pm0.57$} & $16.09${\tiny$\pm0.55$} \\
            TabSyn & $4.23${\tiny$\pm0.35$}  & \redbf{$0.00$}{\tiny\redbf{$\pm0.01$}}  & $8.30${\tiny$\pm0.42$}  & $4.18${\tiny$\pm0.26$}  \\
            GReaT & $1.98${\tiny$\pm0.24$}  & $0.15${\tiny$\pm0.05$}  & $3.51${\tiny$\pm0.28$}  & $1.88${\tiny$\pm0.19$}  \\
            \midrule
            \textbf{GraDe} & \redbf{$0.64$}{\tiny\redbf{$\pm0.14$}}  & $0.08${\tiny$\pm0.04$}  & \redbf{$2.64$}{\tiny\redbf{$\pm0.24$}}  & \redbf{$1.12$}{\tiny\redbf{$\pm0.14$}}  \\
            \textbf{GraDe-Light} & $1.52${\tiny$\pm0.21$}  & $0.13${\tiny$\pm0.04$}  & $3.54${\tiny$\pm0.28$}  & $1.73${\tiny$\pm0.18$} \\
		\bottomrule[1.0pt] 
		\end{tabular}
  }
  \caption{Violation rates of intrinsic constraints in synthetic data (in \%, a score closer to 0\% is better) with 95\% confidence intervals. The best results are marked in \redbf{Red}.} 
  \label{tab:violation}
\end{table*}

\subsection{Intrinsic Constraint Fidelity}
\label{subsec:constraint_fidelity}
For LLM-based data synthesis methods, ensuring synthetic data is both diverse and preserves intrinsic constraints from original datasets is crucial \citep{rao-etal-2024-commonit, rao2025apt, chen2025consistentchat, lu2025mult, qingsong2025raise, han2025att}. 
Following \citet{xu2024llms}, we evaluate how well synthetic data preserves constraints that naturally exist in real-world datasets. Table~\ref{tab:violation} reports violation rates for three representative constraints: geographic coordinates matching state boundaries (Bird), education level matching education code (Income), and housing coordinates falling within California boundaries (Housing).

GraDe achieves the lowest average violation rate with 1.12\%, significantly outperforming conventional generators. This superior performance stems from our explicit modeling of functional dependencies, which captures structural relationships between features. While GReaT also performs well (1.88\% violations), our graph-guided approach provides more consistent constraint preservation. These results confirm that GraDe maintains both statistical properties and factual consistency, essential aspects of high-fidelity synthetic data.

\begin{table*}[ht] 
    \centering
    \renewcommand{\arraystretch}{1.05}
    \footnotesize
    {
        \resizebox{\textwidth}{!}{
	\begin{tabular}{rlc|ccccc|cc}
            \toprule[0.8pt]
            Dataset & & Original & TVAE & CTGAN & CTABGAN+ & TabSyn & GReaT & \textbf{GraDe} & \textbf{GraDe-Light}\\
            \midrule 
            \multirow{3}{*}{Bird ($\uparrow$)} & DT & $99.97${\tiny$\pm0.02$}  & $70.59${\tiny$\pm0.00$} & $94.28${\tiny$\pm0.17$} & $66.46${\tiny$\pm0.21$} & $99.75${\tiny$\pm0.06$}  & $99.63${\tiny$\pm0.07$}  &  $99.76${\tiny$\pm0.02$} & \redbf{$99.78$}{\tiny\redbf{$\pm0.00$}}  \\ 
            & RF & $100.00${\tiny$\pm0.00$} & $79.87${\tiny$\pm0.46$} & $99.09${\tiny$\pm0.08$}  & $75.54${\tiny$\pm0.11$} & $99.97${\tiny$\pm0.12$}  & \redbf{$100.00$}{\tiny\redbf{$\pm0.00$}} & \redbf{$100.00$}{\tiny\redbf{$\pm0.00$}} &  $99.96${\tiny$\pm0.02$}\\
            & LR & $100.00${\tiny$\pm0.00$} & $87.04${\tiny$\pm0.00$} & \redbf{$100.00$}{\tiny\redbf{$\pm0.00$}} & $73.99${\tiny$\pm0.00$} & \redbf{$100.00$}{\tiny\redbf{$\pm0.00$}} & $98.29${\tiny$\pm0.00$}  & \redbf{$100.00$}{\tiny\redbf{$\pm0.00$}} & $98.35${\tiny$\pm0.00$}\\
            \midrule
            
            \multirow{3}{*}{Sick ($\uparrow$)} & DT & $98.70${\tiny$\pm0.10$} & \redbf{$95.87$}{\tiny\redbf{$\pm0.26$}} & $86.28${\tiny$\pm1.20$} & $92.40${\tiny$\pm0.42$} & $74.65${\tiny$\pm1.05$} & $91.42${\tiny$\pm0.61$} & $95.47${\tiny$\pm0.77$} & $92.90${\tiny$\pm0.38$} \\
            & RF & $98.46${\tiny$\pm0.18$} & $95.34${\tiny$\pm0.24$} & $94.97${\tiny$\pm0.00$} & $94.99${\tiny$\pm0.26$} & $96.21${\tiny$\pm0.34$} & $96.40${\tiny$\pm0.15$} & \redbf{$97.77$}{\tiny\redbf{$\pm0.05$}} & $97.51${\tiny$\pm0.27$}\\
            & LR & $89.54${\tiny$\pm0.00$} & \redbf{$94.57$}{\tiny\redbf{$\pm0.00$}} & $58.01${\tiny$\pm0.00$} & $82.65${\tiny$\pm0.00$} & $66.23${\tiny$\pm0.00$} & $83.31${\tiny$\pm0.00$} & $91.52${\tiny$\pm0.00$} & $89.14${\tiny$\pm0.00$}\\
            \midrule   
            
            \multirow{3}{*}{Income ($\uparrow$)} & DT & $81.14${\tiny$\pm0.03$} & $80.28${\tiny$\pm0.12$} & $79.87${\tiny$\pm0.17$} & $76.61${\tiny$\pm0.31$} & $80.73${\tiny$\pm0.13$} & $79.09${\tiny$\pm0.14$} & \redbf{$80.85$}{\tiny\redbf{$\pm0.09$}} & $80.61${\tiny$\pm0.19$} \\
            & RF & $85.15${\tiny$\pm0.15$} & $82.62${\tiny$\pm0.13$} & $82.25${\tiny$\pm0.13$} & $83.38${\tiny$\pm0.18$} & $80.22${\tiny$\pm0.11$} & $83.68${\tiny$\pm0.12$} & \redbf{$84.26$}{\tiny\redbf{$\pm0.10$}} & $84.66${\tiny$\pm0.07$}\\
            & LR & $80.45${\tiny$\pm0.00$} & $78.99${\tiny$\pm0.00$} & $79.19${\tiny$\pm0.00$} & $77.07${\tiny$\pm0.00$} & \redbf{$81.14$}{\tiny\redbf{$\pm0.00$}} & $80.19${\tiny$\pm0.00$} & $80.90${\tiny$\pm0.00$} & $80.13${\tiny$\pm0.00$}\\
            \midrule
         
            \multirow{3}{*}{Diabetes ($\uparrow$)} & DT & $74.68${\tiny$\pm1.30$} & $72.21${\tiny$\pm0.49$} & $59.48${\tiny$\pm0.88$} & $62.99${\tiny$\pm1.23$} & $75.06${\tiny$\pm2.08$} & $65.97${\tiny$\pm0.66$} & \redbf{$76.62$}{\tiny\redbf{$\pm0.88$}} & $72.21${\tiny$\pm1.12$} \\
            & RF & $74.94${\tiny$\pm0.88$} & \redbf{$75.97$}{\tiny\redbf{$\pm0.71$}} & $57.01${\tiny$\pm1.50$} &  $64.81${\tiny$\pm1.50$} & $75.32${\tiny$\pm0.92$} & $67.40${\tiny$\pm0.76$} & $73.90${\tiny$\pm0.95$} & $75.32${\tiny$\pm0.71$} \\
            & LR & $69.48${\tiny$\pm0.00$} & \redbf{$73.38$}{\tiny\redbf{$\pm0.00$}} & $57.14${\tiny$\pm0.00$} & $72.08${\tiny$\pm0.00$} & $70.78${\tiny$\pm0.00$} & $67.53${\tiny$\pm0.00$} & $69.48${\tiny$\pm0.00$} & $68.18${\tiny$\pm0.00$} \\
            \midrule

            \multirow{3}{*}{Housing ($\downarrow$)} & DT & $0.24${\tiny$\pm0.01$} & $0.35${\tiny$\pm0.00$} & $0.50${\tiny$\pm0.00$} & $0.39${\tiny$\pm0.00$} & $0.30${\tiny$\pm0.01$} & $0.34${\tiny$\pm0.00$} & $0.30${\tiny$\pm0.00$} & \redbf{$0.29$}{\tiny\redbf{$\pm0.00$}}  \\
            & RF & $0.18${\tiny$\pm0.00$} & $0.30${\tiny$\pm0.01$} & $0.37${\tiny$\pm0.00$} & $0.28${\tiny$\pm0.01$} & $0.22${\tiny$\pm0.00$} & $0.24${\tiny$\pm0.01$} & \redbf{$0.21$}{\tiny\redbf{$\pm0.01$}}  & $0.22${\tiny$\pm0.00$} \\
            & LR & $0.29${\tiny$\pm0.00$} & $0.33${\tiny$\pm0.00$} & $0.35${\tiny$\pm0.00$} & $0.29${\tiny$\pm0.00$} & \redbf{$0.29$}{\tiny\redbf{$\pm0.00$}} & $0.30${\tiny$\pm0.00$} & $0.30${\tiny$\pm0.00$} & $0.31${\tiny$\pm0.00$} \\
		\bottomrule[1.0pt] 
		\end{tabular}}
  }
  \caption{Additional results of machine learning efficiency experiment. Here, GraDe and GraDe-Light use GPT-2 Medium as backbone model. The best results are marked in \redbf{Red}. Classification accuracy ($\uparrow$) and regression mean absolute percentage error ($\downarrow$) are reported.} 
  \label{tab:mle_gpt2_m}
\end{table*}

\begin{table*}[ht] 
    \centering
    \renewcommand{\arraystretch}{1.1}
    \footnotesize
    {
	\begin{tabular}{lccccc|c}
            \toprule[0.8pt]
            Method & Bird & Sick & Income & Diabetes & Housing & Average \\
            \midrule 
            TVAE & $74.95${\tiny$\pm0.25$} & $76.22${\tiny$\pm2.26$} & $67.27${\tiny$\pm1.21$} & $67.59${\tiny$\pm6.58$} & $60.37${\tiny$\pm0.97$} & $69.28${\tiny$\pm2.26$} \\
            CTGAN & $59.90${\tiny$\pm0.98$} & $81.54${\tiny$\pm2.66$} & $62.58${\tiny$\pm0.81$} & $85.50${\tiny$\pm3.22$} & $62.89${\tiny$\pm0.77$} & $70.48${\tiny$\pm1.69$} \\
            CTABGAN+ & $63.77${\tiny$\pm1.35$} & $64.35${\tiny$\pm2.13$} & \underline{$58.49${\tiny$\pm0.42$}} & $60.75${\tiny$\pm2.47$} & $60.41${\tiny$\pm0.44$} & $61.55${\tiny$\pm1.36$} \\
            TabSyn & \redbf{$49.32$}{\tiny\redbf{$\pm0.37$}} & \redbf{$54.82$}{\tiny\redbf{$\pm2.74$}} & \redbf{$55.41$}{\tiny\redbf{$\pm0.85$}}  & \redbf{$45.36$}{\tiny\redbf{$\pm2.79$}} & \redbf{$50.48$}{\tiny{\redbf{$\pm0.71$}}} & \redbf{$51.08$}{\tiny\redbf{$\pm1.49$}} \\
            GReaT & $53.92${\tiny$\pm0.27$} & $65.48${\tiny$\pm3.85$} & $65.15${\tiny$\pm0.67$} & $68.40${\tiny$\pm3.87$} & $58.14${\tiny$\pm1.35$} & $62.22${\tiny$\pm2.00$} \\
            \midrule
            \textbf{GraDe} & \underline{$52.33${\tiny$\pm0.67$}} & \underline{$62.33${\tiny$\pm1.62$}} & $59.49${\tiny$\pm0.85$} & \underline{$59.28${\tiny$\pm2.37$}} & \underline{$56.54${\tiny$\pm0.80$}} & \underline{$57.99${\tiny$\pm1.26$}} \\
            \textbf{GraDe-Light} & $53.43${\tiny$\pm0.74$} & $63.44${\tiny$\pm2.66$} & $64.64${\tiny$\pm0.90$}  & $66.45${\tiny$\pm2.61$} & $58.40${\tiny$\pm0.47$} & $61.27${\tiny$\pm1.48$} \\
		\bottomrule[1.0pt] 
		\end{tabular}
  }
  \caption{Discriminator measure (accuracy in \%, a score closer to 50\% is better) with 95\% confidence intervals. The best results are marked in \redbf{Red}, second-best results are \underline{underlined}.} 
  \label{tab:discriminator}
  \vspace{-5pt}
\end{table*}

\subsection{Scaling to Larger Backbone Model}
\label{subsec:mle_gpt2_m}
In principle, GraDe can be applied to any decoder-only language models. To investigate the impact of model scale on our approach, we conduct additional experiments using GPT-2 Medium \citep{radford2019language} with 355 million parameters as the backbone model. Table~\ref{tab:mle_gpt2_m} presents the MLE results for both GraDe and GraDe-Light with this larger model. These results complement our main findings and demonstrate the scalability of our approach across different model sizes.

\subsection{Discriminator Measure}
\label{subsec:discriminator}
To further assess the \textit{fidelity} of synthetic data beyond the correlation error histograms reported in \S\ref{sec:results}, we employ a discriminator-based evaluation approach. We train a support vector machine \citep{svm} with 5-fold cross-validation to distinguish between real and synthetic samples. For high-quality synthetic data, the discriminator accuracy should approach 50\% (random chance), indicating the classifier cannot reliably differentiate between real and synthetic distributions.

Table~\ref{tab:discriminator} shows our generated samples effectively confuse classifiers trained on real data. While TabSyn achieves closest to ideal performance at nearly 50\% accuracy, our method outperforms other baselines. TabSyn's advantage likely stems from its score-based latent space guidance directing generation toward high-probability regions that closely mimic the original distribution. However, this close approximation may raise privacy concerns due to significant feature overlap between synthetic and real samples.

\begin{figure}[htbp]
    \centering
    \includegraphics[width=1\linewidth]{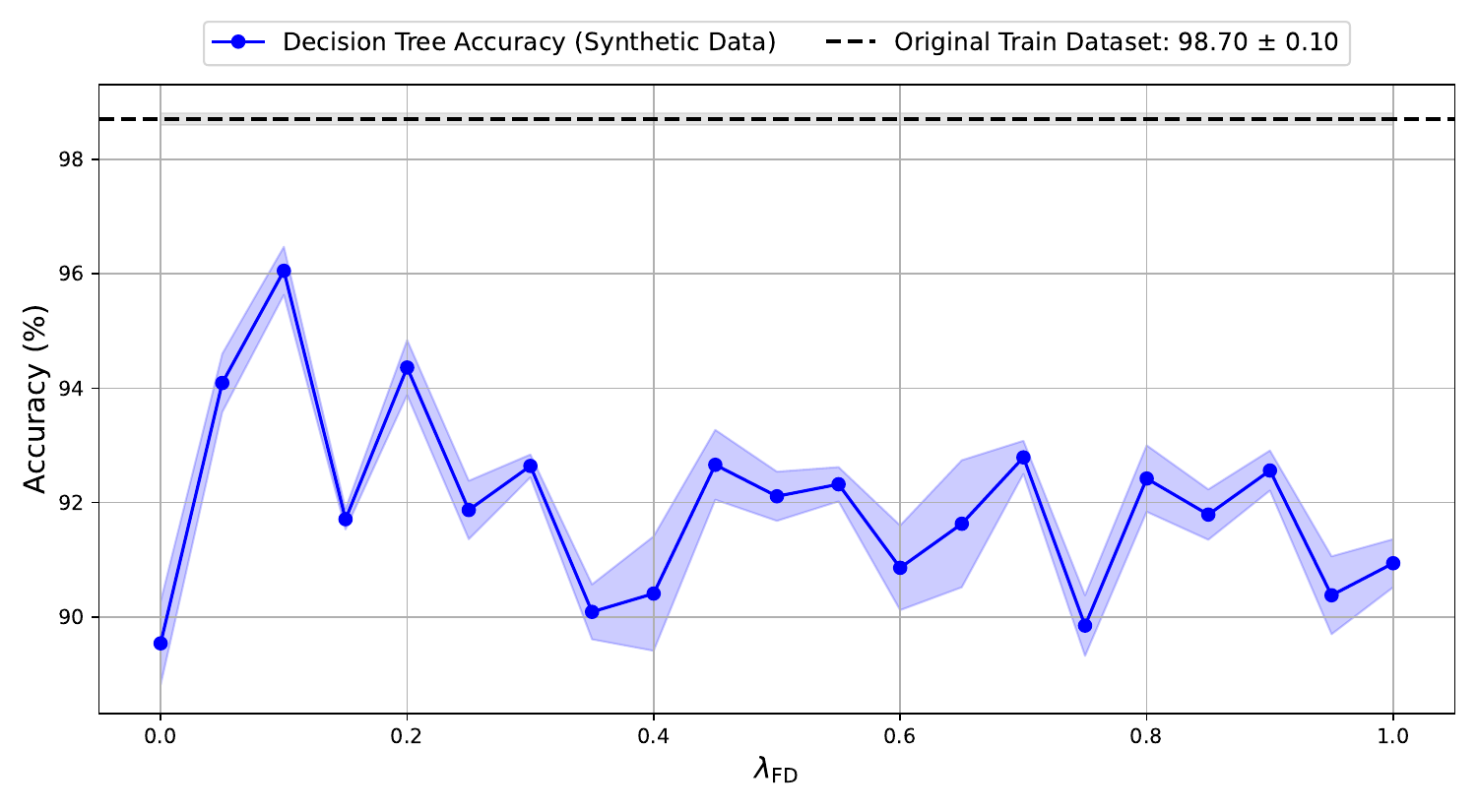}
    \caption{Impact of functional dependency alignment weight $\lambda_{\text{FD}}$ on synthetic data quality, measured by classification accuracy on the \textbf{Sick} dataset.}
    \label{fig:impact_fd}
\end{figure}

\subsection{Impact of FD Alignment Loss}
\label{subsec:impact_fd}

In ablation study we observe that removing FD alignment loss has the most significant impact on medical datasets like Sick (\S\ref{subsec:ablation}). This is likely because such datasets contain many numerous numerical features and semantically ambiguous feature names (such as ``\texttt{T3}'', ``\texttt{TT4}'' in Sick), making it difficult for language models to capture dependencies implicitly without structural guidance. 

To better understand this effect, we investigate how different FD alignment loss weights $\lambda_{\text{FD}}$ (as given in Eq.~\ref{equ:overall_loss}) affect performance on challenging datasets. We train GraDe with varying $\lambda_{\text{FD}}$ values on Sick and evaluate synthetic data quality using decision tree classifiers. Figure~\ref{fig:impact_fd} shows the results, with the dashed line representing models trained on original data. The identified best value for $\lambda_{\text{FD}}$ in Sick is $0.1$.

\subsection{Impact of Sparsity Constraint on High-Dimensional Data}
\label{subsec:ablation_wide}
To further validate the effectiveness of our sparsity constraint in more challenging scenarios, we conducted an ablation study on the UCI Myocardial Infarction Complications (MIC) dataset \citep{giaa128}, which contains 124 features for predicting patient complications. This high-dimensional setting provides a more comprehensive evaluation of the sparsity constraint's impact compared to our initial experiments with 5–30 features.

We generated synthetic data under four different configurations and evaluated performance using decision tree (DT) classifiers, with results averaged across five random seeds. Table~\ref{tab:sparsity_ablation} presents the classification accuracy for each configuration.

\begin{table}[htbp] 
    \centering
    \renewcommand{\arraystretch}{1.1}
    \footnotesize
    {
        \begin{tabular}{lc}
            \toprule[1.0pt]
            Method & DT Accuracy (\%) $\uparrow$ \\
            \midrule
            GraDe (full) & \textbf{89.41} \\
            GraDe w/o Sparsity & 86.18 \\
            GraDe w/o FDs & 87.35 \\
            w/o both & 84.41 \\
		\bottomrule[1.0pt] 
    \end{tabular}
  }
  \caption{Ablation study on sparsity constraint effectiveness in high-dimensional data (124 features). Results show DT accuracy averaged over 5 random seeds.}
  \label{tab:sparsity_ablation}
\end{table}

The results confirm that the sparsity constraint demonstrates notably stronger effectiveness in this higher-dimensional setting, with a performance improvement of 3.23 percentage points compared to the variant without sparsity. Both sparsity and FD constraints prove complementary, with the FD constraint itself acting as an additional form of sparsity by selectively emphasizing essential dependencies. This validates our hypothesis that larger, more complex datasets better showcase the advantages of our constraint-guided approach.

\section{Implementation Details}
\subsection{Models and Evaluation}
\paragraph{Models}
All models are implemented in PyTorch \citep{paszke2019pytorch} with pretrained language models from Huggingface \citep{wolf2019huggingface} as our foundations. We modified their source codes to integrate our graph-guided attention modules. For baselines, we use hyperparameters recommended by their respective authors. All experiments were conducted on an NVIDIA A100 GPU with 80GB memory.

\paragraph{Hyperparameter Settings}
We maintain consistent hyperparameter configurations across all datasets for both GraDe and GraDe-Light. We use the AdamW optimizer \citep{loshchilov2017decoupled} with a learning rate of $5 \times 10^{-5}$. Depending on the GPU memory limitations, we use a fixed batch size of 64. The sparsity and functional dependency regularization weights ($\lambda_{\text{sparse}}$ and $\lambda_{\text{FD}}$) are set to $0.001$ and $0.1$ respectively. For the sampling step, we set the temperature to $0.7$ and the nucleus sampling parameter to $0.95$ for all experiments and datasets. And following \citet{borisov2022language}, we use regular expressions to convert generated text back to a tabular format \citep{10.5555/114872.114877}.

\begin{table}[htbp] 
    \centering
    \renewcommand{\arraystretch}{1.1}
    \footnotesize
    {
        \resizebox{\linewidth}{!}{
        \begin{tabular}{ccccc}
            \toprule
            \textbf{DT} & \multicolumn{2}{c}{\textbf{RF}} & \textbf{LR} & \textbf{SVM} \\
            \cmidrule(lr){1-1} \cmidrule(lr){2-3} \cmidrule(lr){4-4} \cmidrule(lr){5-5}
            max\_depth & max\_depth & n\_estimators & max\_iter & kernel  \\
            \midrule[0.2pt]
            20 & 20 & 100 & 500 & linear \\
            \bottomrule
		\end{tabular}}
  }
  \caption{Hyperparameter settings of the classification and regression models used in our experiments.} 
  \label{tab:mle_para}
\end{table}

\paragraph{Evaluation Settings}
For the machine learning efficiency (\S\ref{sec:results}, \S\ref{subsec:scaling}, \S\ref{subsec:impact_fd} and \S\ref{subsec:ablation_wide}), low-data regimes (\S\ref{subsec:low_data}), ablation study (\S\ref{subsec:ablation}) and discriminator experiments (\S\ref{subsec:discriminator}) we additionally use decision tree (DT), random forest (RF), linear/logistic regression (LR) and support vector machine (SVM) models from the Scikit-Learn package \citep{sklearn_api}. Table~\ref{tab:mle_para} summarizes the hyperparameter settings used for all models in our experiments. For the scaling experiments (\S\ref{subsec:scaling}) on the imbalanced Loan dataset, we additionally enable class weighting for all classifiers to properly account for the class imbalance. 

\subsection{Parameter Counts and Efficiency}
To enhance parameter efficiency, we introduce GraDe-Light, a lightweight variant that updates only the dynamic graph-guided attention modules while freezing all other parameters. As shown in Table~\ref{tab:parameter}, this approach substantially reduces the number of trainable parameters by 77.2\% for GPT-2 and 71.6\% for GPT-2 Medium. Despite this significant reduction, GraDe-Light maintains competitive performance across most datasets, offering a practical solution for computationally constrained environments while preserving the core benefits of our approach.

\begin{table}[htbp] 
    \centering
    \renewcommand{\arraystretch}{1.05}
    \footnotesize
    {
	\resizebox{\linewidth}{!}{\begin{tabular}{l|ccc}
            \toprule[0.8pt]
            Method & Model & \#Parameters & \#Trainable \\
            \midrule
            GraDe & GPT-2 & 124.5M & 124.5M \\
            GraDe-Light & GPT-2 & 124.5M & 28.4M \\
            \midrule
            GraDe & GPT-2 Medium & 355.0M & 355.0M \\
            GraDe-Light & GPT-2 Medium & 355.0M & 100.9M \\
		\bottomrule[1.0pt] 
		\end{tabular}}
  }
  \caption{Parameter comparison between GraDe and GraDe-Light across different foundation models. \#Parameters and \#Trainable denote the number of total parameters and the trainable parameters. Parameter counts are approximate and include the dynamic graph-guided attention modules.} 
  \label{tab:parameter}
\end{table}

\begin{table*}[ht] 
    \centering
    \renewcommand{\arraystretch}{1.2}
    \footnotesize
    \begin{threeparttable}
    \resizebox{\textwidth}{!}{
    \begin{tabular}{lccccccc}
            \toprule[0.8pt]
            Dataset  & Domain & \#Samples & \#Features & Task   & \#Classes &\#Train [Pos\%] &\#Test [Pos\%] \\
            \midrule
            Bird \citep{arcgisStatesShapefile}  & Ecology     & 16,087   & 5     & Classification    & 30  & - & -  \\
            Sick \citep{Dua:2019}  & Medical     & 3772     & 30             & Classification & 2 & 6.4\% & 5.0\%       \\
            Income \citep{Dua:2019}   & Social      & 32,561   & 15         & Classification & 2  & 24.2\% & 23.6\%      \\
            Diabetes (From Kaggle$^{1}$) & Medical     & 768      & 9           & Classification & 2 & 34.7\% & 35.7\%      \\
            Housing \citep{pace1997sparse}  & Real Estate & 20,640   & 10        & Regression     & -  & - & -   \\
            \rowcolor[rgb]{0.95,0.95,0.95}
            Loan (From Kaggle$^{2}$) & Financial    & 252,000   & 13     & Classification & 2  & 12.3\% & 12.4\%      \\
            \rowcolor[rgb]{0.95,0.95,0.95}
            MIC \citep{giaa128} & Medical & 1,700 & 124 & Classification & 2 & 3.16\% & 3.24\%  \\
        \bottomrule[1.0pt] 
        \end{tabular}}
  \begin{tablenotes}
    \item[1] https://www.kaggle.com/datasets/uciml/pima-indians-diabetes-database 
    \item[2] https://www.kaggle.com/datasets/subhamjain/loan-prediction-based-on-customer-behavior
  \end{tablenotes}
  \end{threeparttable}
  \caption{Statistics of the real-world datasets used in our experiments. The first five datasets are used for our main experiments. The shaded Loan and MIC datasets are used for our additional experiments (\S\ref{subsec:scaling}, \S\ref{subsec:ablation_wide}). \#Samples, \#Features and \#Classes denote the numbers of samples, features and classes in tabular datasets, respectively. For binary classification tasks, we report the percentage of positive samples in both training and test sets to highlight class imbalance.} 
  \label{tab:datasets}
\end{table*}

\subsection{Datasets}
\label{subsec:datasets}
We evaluate our approach on six real-world datasets, with five used in our main experiments and two used for our additional experiments. Bellows are the introductions for each dataset:
\begin{itemize}[leftmargin=*, nolistsep]
\vspace{5pt}
\item \textbf{Bird}: The dataset contains geographic coordinates, official state birds, and latitude zones for all 51 US states. The task is to predict the official state bird based on location data.
\vspace{5pt}
\item \textbf{Sick}: The dataset contains patient records with demographic information, medical history, and thyroid hormone measurements. The task is to predict thyroid disease status based on clinical features.
\vspace{5pt}
\item \textbf{Income}: The dataset contains demographic and employment information. The task is to predict whether an individual's income exceeds \$50,000 based on personal attributes.
\vspace{5pt}
\item \textbf{Diabetes}: The dataset collected by the National Institute of Diabetes and Digestive and Kidney Diseases \citep{smith1988using}, contains clinical measurements of women of Pima Indian heritage. The task is to predict whether a patient has diabetes based on health indicators.
\vspace{5pt}
\item \textbf{Housing}: The dataset contains district-level housing and demographic information for California. The task is to predict the median house value of each district.
\vspace{5pt}
\item \textbf{Loan}: The dataset contains personal financial, demographic, and employment information about individuals from various cities in India. The task is to predict the loan risk based on personal attributes.
\vspace{5pt}
\item \textbf{MIC}: The dataset contains clinical measurements from myocardial infarction patients during hospitalization. The task is to predict complications of myocardial infarction based on patient information collected at admission and on the third day of treatment.
\end{itemize}
For all datasets, we use an 80\%/20\% train-test split for model training and evaluation. Table \ref{tab:datasets} provides comprehensive statistics for all datasets.

\subsection{Functional Dependency Extraction}
\label{subsec:fds}
To incorporate functional dependencies as structural priors into GraDe, we employ well-established algorithms from data profiling literature. Specifically, we adopt HyFD~\citep{papenbrock2016hybrid} as our primary extraction method, supplemented by TANE~\citep{huhtala1999tane} as a verification reference.

\paragraph{TANE \citep{huhtala1999tane}:}
A classical, lattice-based FD discovery algorithm that systematically enumerates all attribute combinations using a breadth-first, level-wise traversal. It computes partitions of tuples and exhaustively verifies all candidate dependencies, guaranteeing completeness and soundness. While theoretically rigorous, TANE's computational complexity grows exponentially with the number of attributes, making it impractical for datasets with many features or large-scale data.

\paragraph{HyFD \citep{papenbrock2016hybrid}:}
A more recent, hybrid approach designed explicitly to handle large-scale, real-world datasets efficiently. The algorithm proceeds in two key phases:

In the first phase (the \textit{candidate generation phase}), HyFD employs a sampling-based method that examines subsets of the dataset to rapidly generate a comprehensive list of potential FD candidates. Specifically, it selects random pairs of data tuples, constructs \textit{difference sets} (capturing attribute differences between tuple pairs), and leverages these sets to prune impossible or improbable FD candidates efficiently. This sampling-driven strategy drastically reduces computational cost compared to exhaustive enumeration.

The second phase (the \textit{validation phase}) systematically verifies the previously identified FD candidates by constructing partition-based equivalence classes over the full dataset. HyFD efficiently refines and prunes these partitions to confirm or discard each candidate FD rigorously. This hybrid strategy effectively balances computational efficiency and completeness, supporting both exact and approximate functional dependencies, and ensures high reliability across varied data conditions, including noise or sparsity.

\paragraph{Practical Choice for Structural Guidance}
Given its strong trade-off between completeness and efficiency, we use HyFD throughout our experiments to extract structural priors. Its scalability and robustness to noise make it a practical choice for real-world applications where exhaustive enumeration is infeasible.

\paragraph{Challenges of Automated FD Extraction}
One potential concern with using automatically extracted functional dependencies is the risk that these structures may be incomplete or dataset-specific, raising the question of whether they can reliably guide modeling without manual verification. However, our approach uses extracted FDs not as absolute truth but as helpful structural hints from the data itself. Algorithms like HyFD identify statistically supported patterns that enhance model performance without requiring expert input. GraDe's design acknowledges that these dependencies might be imperfect; during training, the model naturally learns which dependencies are reliable and which are not, adjusting their influence accordingly. This makes our approach practical for real-world applications where manual verification would be prohibitively time-consuming.

\paragraph{Robustness Through Complementary Modeling}
GraDe achieves reliability through two complementary strategies. First, it employs a soft constraint formulation that allows the model to flexibly incorporate dependencies based on their observed reliability. This means GraDe naturally reduces the influence of noisy or incorrect dependencies while benefiting from valid ones. Second, our method combines two sources of structural information: statistical patterns from extracted FDs and semantic understanding from the pretrained LLM. This creates \textit{natural resilience}: when extracted dependencies have flaws, the LLM's language understanding fills the gaps; when language patterns are unclear, the explicit dependencies provide guidance. This balanced approach ensures GraDe works effectively without expert intervention while being robust to imperfections in the dependency extraction process.

We provide the complete lists of FDs identified and utilized in our experiments here:

\input{fds/bird}
\input{fds/sick}
\input{fds/income}
\input{fds/diabetes}
\input{fds/housing}
\input{fds/loan}

\end{document}

%% file: fds/bird.tex
\begin{tcolorbox}[
    colframe=gray!50!black,
    colback=gray!5!white,
    title=\small\texttt{Bird},
    breakable,
    enhanced jigsaw,
]
\scriptsize
\ttfamily
['lon', 'lat'] -> ['state\_code', 'bird']

['lat'] -> ['lat\_zone']

['state\_code'] -> ['bird']
\end{tcolorbox}

%% file: fds/sick.tex
\begin{tcolorbox}[
    colframe=gray!50!black,
    colback=gray!5!white,
    title=\small\texttt{Sick},
    breakable,
    enhanced jigsaw,
]
\scriptsize
\ttfamily
['TSH'] -> ['TSH\_measured', 'TBG\_measured', 'TBG']

['TSH', 'TT4'] -> ['hypopituitary']

['TSH', 'TT4', 'FTI'] -> ['I131\_treatment', 'psych']

['TSH', 'TT4', 'FTI', 'T4U', 'query\_on\_thyroxine'] -> ['Class']

['TSH', 'TT4', 'FTI', 'age'] -> ['T4U']

['TSH', 'TT4', 'FTI', 'tumor', 'age', 'sex'] -> ['referral\_source']

['TSH', 'TT4', 'FTI', 'age', 'T3\_measured', 'sex'] -> ['referral\_source']

['TSH', 'TT4', 'FTI', 'T3'] -> ['thyroid\_surgery', 'lithium']

['TSH', 'TT4', 'FTI', 'T3', 'query\_on\_thyroxine'] -> ['T4U']

['TSH', 'TT4', 'FTI', 'referral\_source'] -> ['thyroid\_surgery', 'lithium']

['TSH', 'TT4', 'FTI', 'T3\_measured', 'referral\_source', 'sex', 'query\_on\_thyroxine'] -> ['T4U']

['TSH', 'TT4', 'FTI', 'referral\_source', 'sex', 'query\_on\_thyroxine', 'Class'] -> ['T4U']

['TSH', 'TT4', 'FTI', 'sex'] -> ['thyroid\_surgery', 'lithium']

['TSH', 'TT4', 'FTI', 'on\_thyroxine'] -> ['thyroid\_surgery']

['TSH', 'TT4', 'FTI', 'T3\_measured'] -> ['thyroid\_surgery', 'Class']

['TSH', 'TT4', 'T4U'] -> ['I131\_treatment', 'lithium']

['TSH', 'TT4', 'T4U', 'age'] -> ['FTI', 'psych', 'FTI\_measured', 'Class']

['TSH', 'TT4', 'T4U', 'age', 'tumor', 'sex'] -> ['referral\_source']

['TSH', 'TT4', 'T4U', 'age', 'T3\_measured', 'sex'] -> ['referral\_source']

['TSH', 'TT4', 'T4U', 'T3'] -> ['query\_on\_thyroxine', 'thyroid\_surgery', 'psych']

['TSH', 'TT4', 'T4U', 'referral\_source'] -> ['psych']

['TSH', 'TT4', 'T4U', 'referral\_source', 'query\_hypothyroid'] -> ['thyroid\_surgery']

['TSH', 'TT4', 'T4U', 'sex'] -> ['thyroid\_surgery']

['TSH', 'TT4', 'T4U', 'sex', 'query\_on\_thyroxine'] -> ['Class']

['TSH', 'TT4', 'T4U', 'sex', 'T3\_measured'] -> ['Class']

['TSH', 'TT4', 'T4U', 'on\_thyroxine', 'query\_hypothyroid'] -> ['thyroid\_surgery']

['TSH', 'TT4', 'T4U', 'query\_on\_thyroxine', 'sick'] -> ['Class']

['TSH', 'TT4', 'T4U', 'sick', 'T3\_measured'] -> ['Class']

['TSH', 'TT4', 'goitre', 'T4U'] -> ['psych']

['TSH', 'TT4', 'T4U', 'T3\_measured'] -> ['query\_on\_thyroxine', 'thyroid\_surgery', 'FTI\_measured']

['TSH', 'TT4', 'T4U', 'Class'] -> ['query\_on\_thyroxine']

['TSH', 'TT4', 'age'] -> ['query\_on\_thyroxine', 'I131\_treatment', 'lithium']

['TSH', 'TT4', 'age', 'T3'] -> ['psych', 'T4U\_measured', 'FTI\_measured']

['TSH', 'TT4', 'age', 'T3', 'sex', 'sick'] -> ['referral\_source']

['TSH', 'TT4', 'age', 'referral\_source'] -> ['psych']

['TSH', 'TT4', 'age', 'referral\_source', 'thyroid\_surgery'] -> ['T4U\_measured', 'FTI\_measured']

['TSH', 'TT4', 'age', 'referral\_source', 'query\_hypothyroid'] -> ['T4U\_measured', 'FTI\_measured']

['TSH', 'TT4', 'age', 'referral\_source', 'Class'] -> ['T4U\_measured', 'FTI\_measured']

['TSH', 'TT4', 'age', 'sex', 'thyroid\_surgery'] -> ['T4U\_measured', 'FTI\_measured']

['TSH', 'TT4', 'age', 'sex', 'query\_hypothyroid'] -> ['T4U\_measured', 'FTI\_measured']

['TSH', 'TT4', 'age', 'sex', 'Class'] -> ['T4U\_measured', 'FTI\_measured']

['TSH', 'TT4', 'age', 'thyroid\_surgery'] -> ['Class']

['TSH', 'TT4', 'age', 'query\_hypothyroid'] -> ['thyroid\_surgery', 'Class']

['TSH', 'TT4', 'age', 'T4U\_measured'] -> ['thyroid\_surgery', 'FTI\_measured', 'Class']

['TSH', 'TT4', 'FTI\_measured', 'age'] -> ['thyroid\_surgery', 'Class']

['TSH', 'TT4', 'Class', 'age'] -> ['thyroid\_surgery']

['TSH', 'TT4', 'T3'] -> ['Class']

['TSH', 'TT4', 'T3', 'referral\_source'] -> ['lithium']

['TSH', 'TT4', 'T3', 'referral\_source', 'sex'] -> ['psych']

['TSH', 'TT4', 'T3', 'referral\_source', 'query\_hypothyroid'] -> ['thyroid\_surgery']

['TSH', 'TT4', 'T3', 'referral\_source', 'tumor'] -> ['thyroid\_surgery']

['TSH', 'TT4', 'T3', 'referral\_source', 'T4U\_measured'] -> ['psych']

['TSH', 'TT4', 'T3', 'referral\_source', 'FTI\_measured'] -> ['psych']

['TSH', 'TT4', 'T3', 'sex'] -> ['thyroid\_surgery', 'I131\_treatment', 'lithium', 'T4U\_measured', 'FTI\_measured']

['TSH', 'TT4', 'T3', 'sex', 'on\_antithyroid\_medication'] -> ['psych']

['TSH', 'TT4', 'T3', 'on\_thyroxine'] -> ['thyroid\_surgery']

['TSH', 'TT4', 'referral\_source', 'sex'] -> ['lithium']

['TSH', 'TT4', 'T3\_measured', 'referral\_source', 'sex', 'query\_hypothyroid'] -> ['thyroid\_surgery']

['TSH', 'TT4', 'T3\_measured', 'referral\_source', 'sex', 'T4U\_measured'] -> ['thyroid\_surgery']

['TSH', 'TT4', 'T3\_measured', 'referral\_source', 'sex', 'FTI\_measured'] -> ['thyroid\_surgery']

['TSH', 'TT4', 'T3\_measured', 'referral\_source', 'sex', 'Class'] -> ['thyroid\_surgery']

['TSH', 'TT4', 'query\_hyperthyroid', 'referral\_source', 'T3\_measured', 'on\_thyroxine', 'T4U\_measured'] -> ['thyroid\_surgery']

['TSH', 'TT4', 'query\_hyperthyroid', 'referral\_source', 'T3\_measured', 'on\_thyroxine', 'FTI\_measured'] -> ['thyroid\_surgery']

['TSH', 'TT4', 'query\_hyperthyroid', 'referral\_source', 'T3\_measured', 'on\_thyroxine', 'Class'] -> ['thyroid\_surgery']

['TSH', 'TT4', 'query\_hyperthyroid', 'referral\_source', 'T3\_measured', 'query\_hypothyroid'] -> ['thyroid\_surgery']

['TSH', 'TT4', 'query\_hyperthyroid', 'tumor', 'referral\_source', 'T3\_measured', 'T4U\_measured'] -> ['thyroid\_surgery']

['TSH', 'TT4', 'query\_hyperthyroid', 'tumor', 'referral\_source', 'T3\_measured', 'FTI\_measured'] -> ['thyroid\_surgery']

['TSH', 'TT4', 'query\_hyperthyroid', 'tumor', 'referral\_source', 'T3\_measured', 'Class'] -> ['thyroid\_surgery']

['TSH', 'TT4', 'psych', 'sex'] -> ['lithium']

['TSH', 'TT4', 'T4U\_measured', 'T3\_measured'] -> ['FTI\_measured']

['TSH', 'TT4', 'FTI\_measured'] -> ['T4U\_measured']

['TSH', 'FTI'] -> ['hypopituitary']

['TSH', 'FTI', 'T4U'] -> ['lithium']

['TSH', 'FTI', 'T4U', 'age'] -> ['query\_on\_thyroxine', 'psych', 'Class']

['TSH', 'FTI', 'T4U', 'age', 'T3', 'TT4\_measured'] -> ['TT4']

['TSH', 'FTI', 'T4U', 'age', 'tumor', 'sex', 'TT4\_measured'] -> ['referral\_source']

['TSH', 'FTI', 'T4U', 'age', 'T3\_measured', 'sex', 'TT4\_measured'] -> ['referral\_source']

['TSH', 'FTI', 'T4U', 'age', 'sex', 'TT4\_measured'] -> ['TT4']

['TSH', 'FTI', 'T4U', 'age', 'on\_thyroxine', 'TT4\_measured'] -> ['TT4']

['TSH', 'FTI', 'T4U', 'age', 'T3\_measured', 'TT4\_measured'] -> ['TT4']

['TSH', 'FTI', 'T4U', 'T3'] -> ['query\_on\_thyroxine', 'thyroid\_surgery', 'psych', 'Class']

['TSH', 'FTI', 'T4U', 'referral\_source'] -> ['psych']

['TSH', 'FTI', 'T4U', 'referral\_source', 'query\_hyperthyroid'] -> ['thyroid\_surgery']

['TSH', 'FTI', 'T4U', 'referral\_source', 'T3\_measured'] -> ['query\_on\_thyroxine']

['TSH', 'FTI', 'T4U', 'referral\_source', 'Class'] -> ['query\_on\_thyroxine']

['TSH', 'FTI', 'T4U', 'sex', 'query\_hyperthyroid'] -> ['thyroid\_surgery']

['TSH', 'FTI', 'T4U', 'on\_thyroxine', 'query\_hyperthyroid'] -> ['thyroid\_surgery']

['TSH', 'FTI', 'T4U', 'query\_hyperthyroid', 'T3\_measured'] -> ['query\_on\_thyroxine', 'thyroid\_surgery']

['TSH', 'FTI', 'T4U', 'query\_hyperthyroid', 'Class'] -> ['query\_on\_thyroxine']

['TSH', 'FTI', 'T4U', 'psych', 'T3\_measured'] -> ['query\_on\_thyroxine']

['TSH', 'FTI', 'T4U', 'psych', 'Class'] -> ['query\_on\_thyroxine']

['TSH', 'FTI', 'age'] -> ['lithium', 'T4U\_measured']

['TSH', 'FTI', 'age', 'T3'] -> ['psych']

['TSH', 'FTI', 'age', 'T3', 'sex'] -> ['referral\_source']

['TSH', 'FTI', 'age', 'T3', 'sex', 'on\_thyroxine'] -> ['T4U']

['TSH', 'FTI', 'age', 'T3', 'sex', 'on\_thyroxine', 'TT4\_measured'] -> ['TT4']

['TSH', 'FTI', 'age', 'T3', 'sex', 'on\_antithyroid\_medication'] -> ['T4U']

['TSH', 'FTI', 'age', 'T3', 'sex', 'TT4\_measured', 'on\_antithyroid\_medication'] -> ['TT4']

['TSH', 'FTI', 'age', 'T3', 'sex', 'query\_hypothyroid'] -> ['T4U']

['TSH', 'FTI', 'age', 'T3', 'sex', 'TT4\_measured', 'query\_hypothyroid'] -> ['TT4']

['TSH', 'query\_hyperthyroid', 'FTI', 'age', 'T3', 'sex'] -> ['T4U']

['TSH', 'query\_hyperthyroid', 'FTI', 'age', 'T3', 'sex', 'TT4\_measured'] -> ['TT4']

['TSH', 'FTI', 'goitre', 'age', 'T3', 'on\_thyroxine'] -> ['T4U']

['TSH', 'FTI', 'goitre', 'age', 'T3', 'on\_thyroxine', 'TT4\_measured'] -> ['TT4']

['TSH', 'FTI', 'goitre', 'age', 'T3', 'on\_antithyroid\_medication'] -> ['T4U']

['TSH', 'FTI', 'goitre', 'age', 'T3', 'TT4\_measured', 'on\_antithyroid\_medication'] -> ['TT4']

['TSH', 'FTI', 'goitre', 'age', 'T3', 'query\_hypothyroid'] -> ['T4U']

['TSH', 'FTI', 'goitre', 'age', 'T3', 'TT4\_measured', 'query\_hypothyroid'] -> ['TT4']

['TSH', 'query\_hyperthyroid', 'FTI', 'goitre', 'age', 'T3'] -> ['T4U']

['TSH', 'query\_hyperthyroid', 'FTI', 'goitre', 'age', 'T3', 'TT4\_measured'] -> ['TT4']

['TSH', 'FTI', 'age', 'referral\_source'] -> ['psych', 'Class']

['TSH', 'FTI', 'age', 'referral\_source', 'sex'] -> ['I131\_treatment']

['TSH', 'FTI', 'age', 'referral\_source', 'on\_thyroxine'] -> ['I131\_treatment']

['TSH', 'FTI', 'age', 'referral\_source', 'TT4\_measured'] -> ['I131\_treatment']

['TSH', 'FTI', 'age', 'sex'] -> ['query\_on\_thyroxine', 'Class']

['TSH', 'FTI', 'age', 'sex', 'on\_thyroxine'] -> ['I131\_treatment']

['TSH', 'FTI', 'age', 'sex', 'T3\_measured'] -> ['psych']

['TSH', 'FTI', 'age', 'on\_thyroxine', 'query\_hypothyroid'] -> ['I131\_treatment']

['TSH', 'FTI', 'age', 'on\_thyroxine', 'query\_hyperthyroid'] -> ['I131\_treatment']

['TSH', 'FTI', 'age', 'pregnant', 'query\_hyperthyroid'] -> ['query\_on\_thyroxine']

['TSH', 'FTI', 'age', 'pregnant', 'tumor'] -> ['query\_on\_thyroxine']

['TSH', 'FTI', 'age', 'T3\_measured'] -> ['query\_on\_thyroxine']

['TSH', 'FTI', 'T3'] -> ['I131\_treatment']

['TSH', 'FTI', 'T3', 'referral\_source'] -> ['lithium', 'Class']

['TSH', 'FTI', 'T3', 'referral\_source', 'on\_thyroxine'] -> ['thyroid\_surgery']

['TSH', 'FTI', 'T3', 'on\_thyroxine', 'sick'] -> ['thyroid\_surgery']

['TSH', 'query\_hyperthyroid', 'FTI', 'T3'] -> ['Class']

['TSH', 'FTI', 'referral\_source', 'thyroid\_surgery', 'query\_hyperthyroid'] -> ['lithium']

['TSH', 'TT4\_measured', 'FTI', 'referral\_source'] -> ['T4U\_measured']

['TSH', 'FTI', 'Class', 'referral\_source'] -> ['lithium']

['TSH', 'TT4\_measured', 'FTI', 'sex'] -> ['T4U\_measured']

['TSH', 'FTI', 'query\_hypothyroid', 'goitre', 'TT4\_measured'] -> ['T4U\_measured']

['TSH', 'FTI', 'goitre', 'TT4\_measured', 'Class'] -> ['T4U\_measured']

['TSH', 'T4U', 'age'] -> ['thyroid\_surgery', 'lithium']

['TSH', 'T4U', 'age', 'T3'] -> ['I131\_treatment']

['TSH', 'T4U', 'age', 'T3', 'sex'] -> ['referral\_source', 'psych']

['TSH', 'T4U', 'age', 'referral\_source'] -> ['FTI\_measured', 'Class']

['TSH', 'T4U', 'age', 'referral\_source', 'sex', 'on\_thyroxine'] -> ['psych']

['TSH', 'T4U', 'age', 'referral\_source', 'sex', 'sick'] -> ['psych']

['TSH', 'query\_hyperthyroid', 'T4U', 'age', 'referral\_source', 'sex'] -> ['psych']

['TSH', 'T4U', 'age', 'sex'] -> ['I131\_treatment']

['TSH', 'T4U', 'age', 'sex', 'on\_thyroxine', 'FTI\_measured'] -> ['psych']

['TSH', 'T4U', 'age', 'sex', 'FTI\_measured', 'sick'] -> ['psych']

['TSH', 'query\_hyperthyroid', 'T4U', 'age', 'sex', 'FTI\_measured'] -> ['psych']

['TSH', 'on\_thyroxine', 'T4U', 'age'] -> ['I131\_treatment']

['TSH', 'T4U', 'age', 'psych'] -> ['FTI\_measured']

['TSH', 'TT4\_measured', 'T4U', 'age'] -> ['I131\_treatment']

['TSH', 'T4U', 'T3'] -> ['FTI\_measured']

['TSH', 'T4U', 'T3', 'referral\_source'] -> ['thyroid\_surgery', 'lithium', 'psych']

['TSH', 'T4U', 'tumor', 'T3', 'referral\_source', 'sex'] -> ['I131\_treatment']

['TSH', 'T4U', 'T3', 'sex'] -> ['Class']

['TSH', 'T4U', 'tumor', 'T3', 'sex', 'on\_thyroxine'] -> ['I131\_treatment']

['TSH', 'T4U', 'T3', 'sex', 'thyroid\_surgery'] -> ['lithium']

['TSH', 'T4U', 'T3', 'on\_thyroxine', 'thyroid\_surgery', 'query\_hypothyroid'] -> ['lithium']

['TSH', 'query\_hyperthyroid', 'T4U', 'T3'] -> ['Class']

['TSH', 'lithium', 'T4U', 'T3'] -> ['thyroid\_surgery']

['TSH', 'T4U', 'referral\_source'] -> ['hypopituitary']

['TSH', 'T4U', 'referral\_source', 'sex', 'T3\_measured'] -> ['FTI\_measured']

['TSH', 'T4U', 'psych', 'referral\_source'] -> ['lithium']

['TSH', 'T4U', 'referral\_source', 'T3\_measured', 'TT4\_measured'] -> ['FTI\_measured']

['TSH', 'T4U', 'T3\_measured', 'psych', 'sex', 'on\_thyroxine', 'thyroid\_surgery', 'query\_hypothyroid'] -> ['lithium']

['TSH', 'T4U', 'T3\_measured', 'psych', 'sex', 'query\_on\_thyroxine'] -> ['FTI\_measured']

['TSH', 'query\_hyperthyroid', 'T4U', 'T3\_measured', 'psych', 'sex'] -> ['FTI\_measured']

['TSH', 'on\_thyroxine', 'T4U'] -> ['hypopituitary']

['TSH', 'query\_on\_thyroxine', 'T4U'] -> ['hypopituitary']

['TSH', 'T4U', 'psych', 'T3\_measured', 'TT4\_measured', 'query\_on\_thyroxine'] -> ['FTI\_measured']

['TSH', 'query\_hyperthyroid', 'T4U', 'psych', 'T3\_measured', 'TT4\_measured'] -> ['FTI\_measured']

['TSH', 'T4U', 'T3\_measured'] -> ['hypopituitary']

['TSH', 'age'] -> ['hypopituitary']

['TSH', 'age', 'T3'] -> ['Class']

['TSH', 'age', 'T3', 'referral\_source'] -> ['lithium', 'psych']

['TSH', 'age', 'T3', 'referral\_source', 'on\_thyroxine', 'query\_hypothyroid'] -> ['thyroid\_surgery']

['TSH', 'age', 'T3', 'referral\_source', 'sick', 'query\_hypothyroid'] -> ['thyroid\_surgery']

['TSH', 'age', 'T3', 'sex'] -> ['thyroid\_surgery', 'lithium']

['TSH', 'FTI\_measured', 'age', 'T3'] -> ['T4U\_measured']

['TSH', 'T4U\_measured', 'age', 'referral\_source'] -> ['FTI\_measured']

['TSH', 'FTI\_measured', 'age', 'referral\_source'] -> ['T4U\_measured']

['TSH', 'FTI\_measured', 'age', 'sex'] -> ['T4U\_measured']

['TSH', 'T4U\_measured', 'age', 'psych'] -> ['FTI\_measured']

['TSH', 'TT4\_measured', 'FTI\_measured', 'age'] -> ['T4U\_measured']

['TSH', 'T3'] -> ['hypopituitary']

['TSH', 'T3', 'TT4\_measured', 'FTI\_measured', 'Class'] -> ['T4U\_measured']

['TSH', 'T4U\_measured', 'T3'] -> ['FTI\_measured']

['TSH', 'referral\_source', 'sex', 'T3\_measured', 'T4U\_measured'] -> ['FTI\_measured']

['TSH', 'referral\_source', 'T3\_measured', 'TT4\_measured', 'T4U\_measured'] -> ['FTI\_measured']

['TSH', 'T3\_measured', 'psych', 'sex', 'query\_on\_thyroxine', 'T4U\_measured'] -> ['FTI\_measured']

['TSH', 'query\_hyperthyroid', 'T3\_measured', 'psych', 'sex', 'T4U\_measured'] -> ['FTI\_measured']

['TSH', 'psych', 'T3\_measured', 'TT4\_measured', 'query\_on\_thyroxine', 'T4U\_measured'] -> ['FTI\_measured']

['TSH', 'query\_on\_thyroxine', 'Class'] -> ['hypopituitary']

['TSH', 'query\_hyperthyroid', 'psych', 'T3\_measured', 'TT4\_measured', 'T4U\_measured'] -> ['FTI\_measured']

['TT4'] -> ['TT4\_measured', 'TBG\_measured', 'TBG']

['TT4', 'FTI'] -> ['hypopituitary', 'T4U\_measured']

['TT4', 'FTI', 'age'] -> ['query\_on\_thyroxine', 'thyroid\_surgery', 'I131\_treatment', 'lithium', 'Class']

['TT4', 'FTI', 'age', 'T3'] -> ['T4U', 'psych']

['TT4', 'FTI', 'age', 'T3', 'referral\_source', 'sex', 'TSH\_measured', 'query\_hypothyroid'] -> ['TSH']

['TT4', 'FTI', 'age', 'T3\_measured', 'referral\_source', 'sex'] -> ['T4U']

['TT4', 'FTI', 'age', 'psych', 'referral\_source', 'T3\_measured'] -> ['T4U']

['TT4', 'FTI', 'age', 'sex'] -> ['psych']

['TT4', 'FTI', 'age', 'T3\_measured', 'sex', 'sick'] -> ['T4U']

['TT4', 'FTI', 'age', 'psych', 'T3\_measured', 'sick'] -> ['T4U']

['TT4', 'FTI', 'T3'] -> ['Class']

['TT4', 'FTI', 'T3', 'referral\_source'] -> ['lithium']

['TT4', 'FTI', 'T3', 'referral\_source', 'sex'] -> ['psych']

['TT4', 'FTI', 'T3', 'TSH\_measured'] -> ['lithium']

['TT4', 'FTI', 'referral\_source', 'sex', 'Class'] -> ['lithium']

['TT4', 'FTI', 'referral\_source', 'on\_thyroxine', 'Class', 'sick', 'I131\_treatment'] -> ['lithium']

['query\_hyperthyroid', 'TT4', 'FTI', 'referral\_source', 'on\_thyroxine', 'Class', 'I131\_treatment'] -> ['lithium']

['TT4', 'FTI', 'referral\_source', 'T3\_measured', 'Class', 'sick', 'I131\_treatment'] -> ['lithium']

['query\_hyperthyroid', 'TT4', 'FTI', 'referral\_source', 'T3\_measured', 'Class', 'I131\_treatment'] -> ['lithium']

['TT4', 'T4U'] -> ['hypopituitary']

['TT4', 'T4U', 'age'] -> ['thyroid\_surgery', 'I131\_treatment', 'lithium']

['TT4', 'T4U', 'age', 'T3'] -> ['FTI', 'query\_on\_thyroxine', 'psych']

['TT4', 'T4U', 'age', 'T3', 'referral\_source', 'sex', 'TSH\_measured', 'query\_hypothyroid'] -> ['TSH']

['TT4', 'T4U', 'age', 'referral\_source'] -> ['psych', 'Class']

['TT4', 'T4U', 'age', 'sex'] -> ['Class']

['on\_thyroxine', 'TT4', 'T4U', 'age'] -> ['Class']

['TT4', 'T4U', 'age', 'query\_hyperthyroid', 'T3\_measured'] -> ['FTI\_measured']

['TT4', 'T4U', 'age', 'TSH\_measured'] -> ['FTI\_measured']

['TT4', 'T4U', 'age', 'T3\_measured'] -> ['query\_on\_thyroxine']

['TT4', 'T4U', 'T3'] -> ['Class']

['TT4', 'T4U', 'T3', 'referral\_source'] -> ['lithium']

['TT4', 'T4U', 'T3', 'referral\_source', 'sex'] -> ['psych']

['TT4', 'T4U', 'T3', 'referral\_source', 'query\_hyperthyroid'] -> ['psych']

['TT4', 'T4U', 'T3', 'sex'] -> ['lithium']

['TT4', 'T4U', 'T3', 'sex', 'on\_thyroxine'] -> ['thyroid\_surgery']

['TT4', 'T4U', 'T3', 'sex', 'query\_hypothyroid'] -> ['thyroid\_surgery']

['TT4', 'T4U', 'T3', 'query\_on\_thyroxine'] -> ['lithium']

['query\_hyperthyroid', 'TT4', 'T4U', 'T3'] -> ['lithium']

['TT4', 'T4U', 'T3', 'TSH\_measured'] -> ['query\_on\_thyroxine', 'lithium']

['TT4', 'T4U', 'psych', 'referral\_source', 'sex', 'Class'] -> ['lithium']

['TT4', 'T4U', 'psych', 'referral\_source', 'on\_thyroxine', 'Class', 'sick', 'I131\_treatment'] -> ['lithium']

['query\_hyperthyroid', 'TT4', 'T4U', 'psych', 'referral\_source', 'on\_thyroxine', 'Class', 'I131\_treatment'] -> ['lithium']

['TT4', 'T4U', 'psych', 'referral\_source', 'T3\_measured', 'Class', 'sick', 'I131\_treatment'] -> ['lithium']

['query\_hyperthyroid', 'TT4', 'T4U', 'psych', 'referral\_source', 'T3\_measured', 'Class', 'I131\_treatment'] -> ['lithium']

['TT4', 'age'] -> ['hypopituitary']

['TT4', 'age', 'T3'] -> ['thyroid\_surgery', 'Class']

['TT4', 'age', 'T3', 'referral\_source'] -> ['psych']

['TT4', 'age', 'T3', 'referral\_source', 'sex'] -> ['I131\_treatment']

['TT4', 'age', 'T3', 'referral\_source', 'sex', 'TSH\_measured'] -> ['T4U\_measured', 'FTI\_measured']

['TT4', 'age', 'T3', 'sex'] -> ['query\_on\_thyroxine']

['TT4', 'age', 'T3', 'sex', 'on\_thyroxine'] -> ['I131\_treatment']

['TT4', 'age', 'T3', 'sex', 'T4U\_measured'] -> ['I131\_treatment']

['TT4', 'age', 'T3', 'sex', 'FTI\_measured'] -> ['I131\_treatment']

['TT4', 'age', 'T3', 'query\_hypothyroid'] -> ['query\_on\_thyroxine']

['TT4', 'age', 'T3', 'TSH\_measured'] -> ['query\_on\_thyroxine']

['TT4', 'age', 'T3\_measured', 'referral\_source', 'sex', 'on\_thyroxine', 'thyroid\_surgery'] -> ['I131\_treatment']

['TT4', 'age', 'query\_hyperthyroid', 'T3\_measured', 'T4U\_measured'] -> ['FTI\_measured']

['TT4', 'age', 'TSH\_measured', 'T4U\_measured'] -> ['FTI\_measured']

['TT4', 'FTI\_measured', 'age'] -> ['T4U\_measured']

['TT4', 'T3'] -> ['hypopituitary']

['TT4', 'FTI\_measured', 'T3'] -> ['T4U\_measured']

['TT4', 'query\_on\_thyroxine'] -> ['hypopituitary']

['FTI'] -> ['FTI\_measured', 'TBG\_measured', 'TBG']

['FTI', 'T4U'] -> ['hypopituitary']

['FTI', 'T4U', 'age'] -> ['lithium']

['FTI', 'T4U', 'age', 'T3'] -> ['I131\_treatment']

['FTI', 'T4U', 'age', 'T3', 'referral\_source', 'sex', 'on\_thyroxine', 'query\_hypothyroid', 'TSH\_measured'] -> ['TSH']

['FTI', 'T4U', 'age', 'T3', 'referral\_source', 'sex', 'on\_thyroxine', 'TT4\_measured'] -> ['TT4']

['FTI', 'T4U', 'age', 'referral\_source', 'sex'] -> ['psych', 'Class']

['FTI', 'T4U', 'age', 'referral\_source', 'sick'] -> ['psych', 'Class']

['FTI', 'T4U', 'age', 'referral\_source', 'query\_hypothyroid'] -> ['psych', 'Class']

['FTI', 'T4U', 'age', 'referral\_source', 'query\_hyperthyroid'] -> ['query\_on\_thyroxine']

['FTI', 'T4U', 'age', 'referral\_source', 'psych'] -> ['Class']

['FTI', 'T4U', 'age', 'referral\_source', 'T3\_measured'] -> ['psych', 'Class']

['FTI', 'T4U', 'age', 'referral\_source', 'TT4\_measured'] -> ['psych', 'Class']

['FTI', 'T4U', 'age', 'referral\_source', 'Class'] -> ['psych']

['FTI', 'T4U', 'age', 'sex', 'query\_hyperthyroid'] -> ['query\_on\_thyroxine']

['on\_thyroxine', 'FTI', 'T4U', 'age'] -> ['I131\_treatment']

['FTI', 'T4U', 'age', 'query\_hyperthyroid', 'psych'] -> ['query\_on\_thyroxine']

['FTI', 'T4U', 'age', 'query\_hyperthyroid', 'TSH\_measured'] -> ['query\_on\_thyroxine']

['FTI', 'T4U', 'age', 'query\_hyperthyroid', 'T3\_measured'] -> ['query\_on\_thyroxine']

['FTI', 'T4U', 'T3', 'referral\_source'] -> ['lithium']

['FTI', 'T4U', 'T3', 'referral\_source', 'sex', 'sick'] -> ['psych']

['FTI', 'T4U', 'T3', 'referral\_source', 'sex', 'query\_hypothyroid'] -> ['psych']

['FTI', 'T4U', 'T3', 'referral\_source', 'sex', 'Class'] -> ['psych']

['FTI', 'T4U', 'T3', 'referral\_source', 'sick'] -> ['Class']

['query\_hyperthyroid', 'FTI', 'T4U', 'T3', 'referral\_source', 'sick'] -> ['psych']

['FTI', 'T4U', 'T3', 'referral\_source', 'query\_hypothyroid'] -> ['Class']

['query\_hyperthyroid', 'FTI', 'T4U', 'T3', 'referral\_source', 'query\_hypothyroid'] -> ['psych']

['query\_hyperthyroid', 'FTI', 'T4U', 'T3', 'referral\_source', 'Class'] -> ['psych']

['FTI', 'T4U', 'T3', 'referral\_source', 'psych'] -> ['Class']

['FTI', 'T4U', 'T3', 'sex'] -> ['lithium']

['FTI', 'T4U', 'T3', 'sex', 'TSH\_measured'] -> ['query\_on\_thyroxine']

['query\_on\_thyroxine', 'FTI', 'T4U', 'T3'] -> ['lithium']

['query\_hyperthyroid', 'FTI', 'T4U', 'T3'] -> ['lithium']

['FTI', 'T4U', 'T3', 'query\_hyperthyroid', 'TSH\_measured'] -> ['query\_on\_thyroxine']

['FTI', 'T4U', 'T3', 'TSH\_measured'] -> ['lithium']

['query\_hyperthyroid', 'FTI', 'T4U', 'psych', 'referral\_source', 'sex', 'Class'] -> ['lithium']

['query\_hyperthyroid', 'FTI', 'T4U', 'psych', 'referral\_source', 'on\_thyroxine', 'Class', 'I131\_treatment'] -> ['lithium']

['query\_hyperthyroid', 'FTI', 'T4U', 'psych', 'referral\_source', 'T3\_measured', 'Class', 'I131\_treatment'] -> ['lithium']

['FTI', 'age'] -> ['hypopituitary']

['FTI', 'age', 'T3'] -> ['query\_on\_thyroxine', 'thyroid\_surgery', 'T4U\_measured', 'Class']

['FTI', 'age', 'T3', 'referral\_source'] -> ['I131\_treatment']

['FTI', 'age', 'T3', 'referral\_source', 'psych'] -> ['lithium']

['FTI', 'age', 'T3', 'sex'] -> ['I131\_treatment']

['FTI', 'age', 'psych', 'referral\_source', 'sex', 'T3\_measured'] -> ['lithium']

['query\_hyperthyroid', 'FTI', 'age', 'psych', 'referral\_source', 'T3\_measured'] -> ['lithium']

['FTI', 'sick', 'age', 'T3\_measured'] -> ['T4U\_measured']

['TT4\_measured', 'FTI', 'sick', 'age'] -> ['T4U\_measured']

['FTI', 'Class', 'age', 'T3\_measured'] -> ['T4U\_measured']

['TT4\_measured', 'FTI', 'Class', 'age'] -> ['T4U\_measured']

['FTI', 'T3'] -> ['hypopituitary']

['TT4\_measured', 'FTI', 'T3'] -> ['T4U\_measured']

['FTI', 'referral\_source', 'sex'] -> ['hypopituitary']

['on\_thyroxine', 'FTI', 'tumor', 'referral\_source'] -> ['hypopituitary']

['FTI', 'Class', 'sex'] -> ['hypopituitary']

['on\_thyroxine', 'FTI', 'Class'] -> ['hypopituitary']

['query\_on\_thyroxine', 'FTI'] -> ['hypopituitary']

['T4U'] -> ['T4U\_measured', 'TBG\_measured', 'TBG']

['T4U', 'age', 'T3'] -> ['FTI\_measured']

['T4U', 'age', 'T3', 'referral\_source'] -> ['lithium', 'psych']

['T4U', 'age', 'T3', 'sex'] -> ['lithium', 'Class']

['T4U', 'age', 'T3', 'sex', 'TSH\_measured'] -> ['I131\_treatment']

['T4U', 'age', 'T3', 'query\_hypothyroid'] -> ['Class']

['T4U', 'age', 'referral\_source'] -> ['hypopituitary']

['T4U', 'age', 'psych', 'referral\_source', 'sex', 'query\_hypothyroid'] -> ['lithium']

['query\_hyperthyroid', 'T4U', 'age', 'T3\_measured', 'referral\_source', 'sex'] -> ['FTI\_measured']

['T4U', 'age', 'referral\_source', 'sex', 'TSH\_measured'] -> ['FTI\_measured']

['query\_hyperthyroid', 'T4U', 'age', 'referral\_source', 'T3\_measured', 'TT4\_measured'] -> ['FTI\_measured']

['T4U', 'age', 'referral\_source', 'TSH\_measured', 'TT4\_measured'] -> ['FTI\_measured']

['T4U', 'age', 'sex'] -> ['hypopituitary']

['query\_hyperthyroid', 'T4U', 'age', 'psych', 'T3\_measured', 'sex'] -> ['FTI\_measured']

['T4U', 'age', 'sex', 'psych', 'TSH\_measured'] -> ['FTI\_measured']

['query\_on\_thyroxine', 'T4U', 'age'] -> ['hypopituitary']

['T4U', 'age', 'TSH\_measured'] -> ['hypopituitary']

['T4U', 'age', 'T3\_measured'] -> ['hypopituitary']

['T4U', 'T3', 'referral\_source', 'query\_hypothyroid'] -> ['hypopituitary']

['T4U', 'T3', 'sex'] -> ['hypopituitary']

['query\_on\_thyroxine', 'T4U', 'T3'] -> ['hypopituitary']

['query\_on\_thyroxine', 'T4U', 'referral\_source'] -> ['hypopituitary']

['query\_on\_thyroxine', 'T4U', 'TSH\_measured'] -> ['hypopituitary']

['Class', 'T4U'] -> ['hypopituitary']

['age'] -> ['TBG\_measured', 'TBG']

['age', 'T3'] -> ['hypopituitary']

['age', 'T3', 'referral\_source', 'on\_thyroxine', 'FTI\_measured', 'thyroid\_surgery'] -> ['T4U\_measured']

['age', 'T3', 'sex', 'on\_thyroxine', 'FTI\_measured', 'thyroid\_surgery'] -> ['T4U\_measured']

['TT4\_measured', 'FTI\_measured', 'age', 'T3'] -> ['T4U\_measured']

['T4U\_measured', 'age', 'T3'] -> ['FTI\_measured']

['age', 'referral\_source', 'sex'] -> ['hypopituitary']

['age', 'referral\_source', 'sex', 'TT4\_measured', 'FTI\_measured', 'sick'] -> ['T4U\_measured']

['query\_hyperthyroid', 'age', 'T3\_measured', 'referral\_source', 'sex', 'T4U\_measured'] -> ['FTI\_measured']

['age', 'referral\_source', 'sex', 'TSH\_measured', 'T4U\_measured'] -> ['FTI\_measured']

['age', 'referral\_source', 'sex', 'TT4\_measured', 'FTI\_measured', 'Class'] -> ['T4U\_measured']

['query\_on\_thyroxine', 'age', 'referral\_source'] -> ['hypopituitary']

['query\_hyperthyroid', 'age', 'referral\_source', 'T3\_measured', 'TT4\_measured', 'T4U\_measured'] -> ['FTI\_measured']

['age', 'referral\_source', 'TSH\_measured', 'TT4\_measured', 'T4U\_measured'] -> ['FTI\_measured']

['query\_on\_thyroxine', 'age', 'sex'] -> ['hypopituitary']

['query\_hyperthyroid', 'age', 'psych', 'T3\_measured', 'sex', 'T4U\_measured'] -> ['FTI\_measured']

['age', 'sex', 'psych', 'TSH\_measured', 'T4U\_measured'] -> ['FTI\_measured']

['query\_on\_thyroxine', 'age', 'TSH\_measured'] -> ['hypopituitary']

['Class', 'age'] -> ['hypopituitary']

['T3'] -> ['T3\_measured', 'TBG\_measured', 'TBG']

['Class', 'T3'] -> ['hypopituitary']

['referral\_source'] -> ['TBG\_measured', 'TBG']

['referral\_source', 'sex', 'query\_on\_thyroxine', 'goitre', 'Class'] -> ['hypopituitary']

['referral\_source', 'query\_on\_thyroxine', 'sick', 'goitre', 'Class'] -> ['hypopituitary']

['sex'] -> ['TBG\_measured', 'TBG']

['sex', 'query\_on\_thyroxine', 'goitre', 'TSH\_measured', 'Class'] -> ['hypopituitary']

['on\_thyroxine'] -> ['TBG\_measured', 'TBG']

['query\_on\_thyroxine'] -> ['TBG\_measured', 'TBG']

['query\_on\_thyroxine', 'sick', 'goitre', 'TSH\_measured', 'Class'] -> ['hypopituitary']

['on\_antithyroid\_medication'] -> ['TBG\_measured', 'TBG']

['sick'] -> ['TBG\_measured', 'TBG']

['pregnant'] -> ['TBG\_measured', 'TBG']

['thyroid\_surgery'] -> ['TBG\_measured', 'TBG']

['I131\_treatment'] -> ['TBG\_measured', 'TBG']

['query\_hypothyroid'] -> ['TBG\_measured', 'TBG']

['query\_hyperthyroid'] -> ['TBG\_measured', 'TBG']

['lithium'] -> ['TBG\_measured', 'TBG']

['goitre'] -> ['TBG\_measured', 'TBG']

['tumor'] -> ['TBG\_measured', 'TBG']

['hypopituitary'] -> ['TBG\_measured', 'TBG']

['psych'] -> ['TBG\_measured', 'TBG']

['TSH\_measured'] -> ['TBG\_measured', 'TBG']

['T3\_measured'] -> ['TBG\_measured', 'TBG']

['TT4\_measured'] -> ['TBG\_measured', 'TBG']

['T4U\_measured'] -> ['TBG\_measured', 'TBG']

['FTI\_measured'] -> ['TBG\_measured', 'TBG']

['Class'] -> ['TBG\_measured', 'TBG']

['TBG\_measured'] -> ['TBG']

['TBG'] -> ['TBG\_measured']
\end{tcolorbox}

%% file: fds/income.tex
\begin{tcolorbox}[
    colframe=gray!50!black,
    colback=gray!5!white,
    title=\small\texttt{Income},
    breakable,
    enhanced jigsaw,
]
\scriptsize
\ttfamily
['fnlwgt', 'capital-gain', 'hours-per-week', 'capital-loss', 'education', 'workclass', 'relationship'] -> ['gender']

['fnlwgt', 'capital-gain', 'hours-per-week', 'capital-loss', 'education-num', 'workclass', 'relationship'] -> ['gender']

['fnlwgt', 'capital-gain', 'hours-per-week', 'age', 'education', 'occupation', 'workclass', 'relationship'] -> ['income']

['fnlwgt', 'capital-gain', 'hours-per-week', 'age', 'education-num', 'occupation', 'workclass', 'relationship'] -> ['income']

['fnlwgt', 'capital-gain', 'hours-per-week', 'age', 'occupation', 'relationship'] -> ['marital-status']

['fnlwgt', 'capital-gain', 'hours-per-week', 'education', 'workclass', 'relationship', 'race'] -> ['gender']

['fnlwgt', 'capital-gain', 'hours-per-week', 'education', 'workclass', 'relationship', 'income'] -> ['gender']

['fnlwgt', 'capital-gain', 'hours-per-week', 'education-num', 'workclass', 'relationship', 'race'] -> ['gender']

['fnlwgt', 'capital-gain', 'hours-per-week', 'education-num', 'workclass', 'relationship', 'income'] -> ['gender']

['fnlwgt', 'hours-per-week', 'age', 'education'] -> ['race']

['fnlwgt', 'hours-per-week', 'age', 'education', 'occupation', 'workclass'] -> ['capital-loss']

['fnlwgt', 'hours-per-week', 'age', 'education', 'occupation', 'relationship'] -> ['marital-status']

['fnlwgt', 'hours-per-week', 'age', 'education-num'] -> ['race']

['fnlwgt', 'hours-per-week', 'age', 'education-num', 'occupation', 'workclass'] -> ['capital-loss']

['fnlwgt', 'hours-per-week', 'age', 'education-num', 'occupation', 'relationship'] -> ['marital-status']

['fnlwgt', 'hours-per-week', 'age', 'occupation', 'workclass'] -> ['race']

['fnlwgt', 'hours-per-week', 'age', 'workclass', 'marital-status', 'relationship'] -> ['race']

['fnlwgt', 'hours-per-week', 'native-country', 'education', 'marital-status', 'relationship'] -> ['race']

['fnlwgt', 'hours-per-week', 'native-country', 'education-num', 'marital-status', 'relationship'] -> ['race']

['fnlwgt', 'hours-per-week', 'native-country', 'occupation', 'workclass', 'gender'] -> ['race']

['fnlwgt', 'hours-per-week', 'native-country', 'occupation', 'marital-status'] -> ['race']

['fnlwgt', 'hours-per-week', 'native-country', 'occupation', 'relationship'] -> ['race']

['fnlwgt', 'hours-per-week', 'native-country', 'occupation', 'gender', 'income'] -> ['race']

['fnlwgt', 'hours-per-week', 'education', 'marital-status', 'relationship', 'income'] -> ['race']

['fnlwgt', 'hours-per-week', 'education-num', 'marital-status', 'relationship', 'income'] -> ['race']

['fnlwgt', 'occupation', 'hours-per-week', 'relationship'] -> ['gender']

['fnlwgt', 'hours-per-week', 'workclass', 'marital-status', 'relationship'] -> ['gender']

['fnlwgt', 'age', 'native-country', 'education'] -> ['race']

['fnlwgt', 'age', 'native-country', 'education-num'] -> ['race']

['fnlwgt', 'age', 'native-country', 'occupation', 'workclass'] -> ['race']

['fnlwgt', 'age', 'native-country', 'workclass', 'marital-status'] -> ['race']

['fnlwgt', 'marital-status', 'age', 'education'] -> ['race', 'gender']

['fnlwgt', 'relationship', 'age', 'education'] -> ['race']

['fnlwgt', 'age', 'education', 'income'] -> ['race']

['fnlwgt', 'marital-status', 'age', 'education-num'] -> ['race', 'gender']

['fnlwgt', 'relationship', 'age', 'education-num'] -> ['race']

['fnlwgt', 'age', 'income', 'education-num'] -> ['race']

['fnlwgt', 'occupation', 'marital-status', 'age'] -> ['race', 'gender']

['fnlwgt', 'occupation', 'relationship', 'age'] -> ['race']

['fnlwgt', 'occupation', 'age', 'income'] -> ['race']

['fnlwgt', 'age', 'workclass', 'marital-status', 'relationship', 'income'] -> ['race']

['fnlwgt', 'relationship', 'age'] -> ['gender']

['fnlwgt', 'native-country', 'education', 'occupation', 'gender'] -> ['race']

['fnlwgt', 'native-country', 'education', 'workclass', 'marital-status', 'relationship', 'gender'] -> ['race']

['fnlwgt', 'native-country', 'education-num', 'occupation', 'gender'] -> ['race']

['fnlwgt', 'native-country', 'education-num', 'workclass', 'marital-status', 'relationship', 'gender'] -> ['race']

['fnlwgt', 'education', 'occupation', 'workclass', 'relationship'] -> ['gender']

['fnlwgt', 'occupation', 'relationship', 'education'] -> ['race']

['fnlwgt', 'education', 'occupation', 'relationship', 'income'] -> ['gender']

['fnlwgt', 'education', 'workclass', 'marital-status', 'relationship', 'gender', 'income'] -> ['race']

['fnlwgt', 'education-num', 'occupation', 'workclass', 'relationship'] -> ['gender']

['fnlwgt', 'occupation', 'relationship', 'education-num'] -> ['race']

['fnlwgt', 'education-num', 'occupation', 'relationship', 'income'] -> ['gender']

['fnlwgt', 'education-num', 'workclass', 'marital-status', 'relationship', 'gender', 'income'] -> ['race']

['fnlwgt', 'occupation', 'marital-status', 'relationship'] -> ['gender']

['education'] -> ['education-num']

['education-num'] -> ['education']
\end{tcolorbox}

%% file: fds/diabetes.tex
\begin{tcolorbox}[
    colframe=gray!50!black,
    colback=gray!5!white,
    title=\small\texttt{Diabetes},
    breakable,
    enhanced jigsaw,
]
\scriptsize
\ttfamily
['DiabetesPedigreeFunction', 'BMI', 'Insulin'] -> ['Glucose', 'Age', 'SkinThickness', 'BloodPressure', 'Pregnancies', 'Diabetes']

['DiabetesPedigreeFunction', 'BMI', 'Glucose'] -> ['Insulin', 'Age', 'SkinThickness', 'BloodPressure', 'Pregnancies', 'Diabetes']

['DiabetesPedigreeFunction', 'BMI', 'Age'] -> ['Insulin', 'Glucose', 'SkinThickness', 'BloodPressure', 'Pregnancies', 'Diabetes']

['DiabetesPedigreeFunction', 'BMI', 'SkinThickness'] -> ['Insulin', 'Glucose', 'Age', 'BloodPressure', 'Pregnancies', 'Diabetes']

['DiabetesPedigreeFunction', 'BMI', 'BloodPressure'] -> ['Insulin', 'Glucose', 'Age', 'SkinThickness', 'Pregnancies', 'Diabetes']

['DiabetesPedigreeFunction', 'BMI', 'Pregnancies'] -> ['Insulin', 'Glucose', 'Age', 'SkinThickness', 'BloodPressure', 'Diabetes']

['DiabetesPedigreeFunction', 'BMI', 'Diabetes'] -> ['Insulin', 'Glucose', 'Age', 'SkinThickness', 'BloodPressure', 'Pregnancies']

['DiabetesPedigreeFunction', 'Insulin', 'Glucose'] -> ['BMI', 'Age', 'SkinThickness', 'BloodPressure', 'Pregnancies', 'Diabetes']

['DiabetesPedigreeFunction', 'Insulin', 'Age', 'Pregnancies'] -> ['SkinThickness']

['DiabetesPedigreeFunction', 'Insulin', 'BloodPressure'] -> ['SkinThickness', 'Diabetes']

['DiabetesPedigreeFunction', 'Glucose', 'Age'] -> ['BMI', 'Insulin', 'SkinThickness', 'BloodPressure', 'Pregnancies', 'Diabetes']

['DiabetesPedigreeFunction', 'Glucose', 'SkinThickness'] -> ['BMI', 'Insulin', 'Age', 'BloodPressure', 'Pregnancies', 'Diabetes']

['DiabetesPedigreeFunction', 'Glucose', 'BloodPressure'] -> ['Diabetes']

['DiabetesPedigreeFunction', 'Glucose', 'Pregnancies'] -> ['BMI', 'Insulin', 'Age', 'SkinThickness', 'BloodPressure', 'Diabetes']

['DiabetesPedigreeFunction', 'Diabetes', 'Glucose'] -> ['BloodPressure']

['DiabetesPedigreeFunction', 'Age', 'SkinThickness'] -> ['Insulin']

['DiabetesPedigreeFunction', 'Age', 'BloodPressure'] -> ['BMI', 'Insulin', 'Glucose', 'SkinThickness', 'Pregnancies', 'Diabetes']

['DiabetesPedigreeFunction', 'Age', 'Pregnancies'] -> ['Diabetes']

['DiabetesPedigreeFunction', 'SkinThickness', 'BloodPressure'] -> ['Insulin', 'Diabetes']

['DiabetesPedigreeFunction', 'SkinThickness', 'Pregnancies'] -> ['Insulin', 'Diabetes']

['DiabetesPedigreeFunction', 'BloodPressure', 'Pregnancies'] -> ['Insulin', 'SkinThickness', 'Diabetes']

['BMI', 'Insulin', 'Glucose', 'Age'] -> ['DiabetesPedigreeFunction', 'SkinThickness', 'BloodPressure']

['BMI', 'Insulin', 'Glucose', 'BloodPressure'] -> ['DiabetesPedigreeFunction', 'Age', 'SkinThickness', 'Pregnancies']

['BMI', 'Insulin', 'Glucose', 'Pregnancies'] -> ['DiabetesPedigreeFunction', 'Age', 'SkinThickness', 'BloodPressure', 'Diabetes']

['BMI', 'Insulin', 'Age', 'SkinThickness', 'Pregnancies'] -> ['DiabetesPedigreeFunction', 'Glucose', 'BloodPressure']

['BMI', 'Glucose', 'Age'] -> ['Pregnancies', 'Diabetes']

['BMI', 'Glucose', 'Age', 'SkinThickness'] -> ['DiabetesPedigreeFunction', 'BloodPressure']

['BMI', 'Glucose', 'Age', 'BloodPressure'] -> ['DiabetesPedigreeFunction', 'Insulin', 'SkinThickness']

['BMI', 'Glucose', 'SkinThickness'] -> ['Insulin']

['BMI', 'Glucose', 'SkinThickness', 'BloodPressure'] -> ['DiabetesPedigreeFunction', 'Age', 'Pregnancies']

['BMI', 'Glucose', 'SkinThickness', 'Pregnancies'] -> ['DiabetesPedigreeFunction', 'Age', 'BloodPressure', 'Diabetes']

['Diabetes', 'BMI', 'Glucose', 'SkinThickness'] -> ['DiabetesPedigreeFunction', 'Age', 'BloodPressure', 'Pregnancies']

['BMI', 'Glucose', 'BloodPressure'] -> ['Diabetes']

['BMI', 'Glucose', 'BloodPressure', 'Pregnancies'] -> ['DiabetesPedigreeFunction', 'Insulin', 'Age', 'SkinThickness']

['Diabetes', 'BMI', 'Glucose', 'Pregnancies'] -> ['Age']

['BMI', 'Age', 'SkinThickness', 'BloodPressure'] -> ['Insulin', 'Diabetes']

['BMI', 'Age', 'SkinThickness', 'BloodPressure', 'Pregnancies'] -> ['DiabetesPedigreeFunction', 'Glucose']

['BMI', 'Age', 'SkinThickness', 'Pregnancies'] -> ['Diabetes']

['BMI', 'SkinThickness', 'BloodPressure', 'Pregnancies'] -> ['Insulin']

['Insulin', 'Glucose', 'Age', 'BloodPressure'] -> ['DiabetesPedigreeFunction', 'BMI', 'SkinThickness', 'Pregnancies', 'Diabetes']

['Insulin', 'Glucose', 'Age', 'Pregnancies'] -> ['Diabetes']

['Glucose', 'Age', 'SkinThickness', 'BloodPressure'] -> ['DiabetesPedigreeFunction', 'BMI', 'Insulin', 'Pregnancies', 'Diabetes']

['Glucose', 'Age', 'SkinThickness', 'Pregnancies'] -> ['Diabetes']

['Glucose', 'Age', 'BloodPressure', 'Pregnancies'] -> ['DiabetesPedigreeFunction', 'BMI', 'Insulin', 'SkinThickness', 'Diabetes']

['Glucose', 'SkinThickness', 'BloodPressure', 'Pregnancies'] -> ['Insulin']

['Glucose', 'SkinThickness', 'BloodPressure', 'Pregnancies', 'Diabetes'] -> ['DiabetesPedigreeFunction', 'BMI', 'Age']
\end{tcolorbox}

%% file: fds/housing.tex
\begin{tcolorbox}[
    colframe=gray!50!black,
    colback=gray!5!white,
    title=\small\texttt{Housing},
    breakable,
    enhanced jigsaw,
]
\scriptsize
\ttfamily
['median\_income', 'total\_rooms', 'median\_house\_value'] -> ['population', 'total\_bedrooms', 'households', 'latitude', 'longitude', 'housing\_median\_age', 'ocean\_proximity']

['median\_income', 'total\_rooms', 'population'] -> ['median\_house\_value', 'total\_bedrooms', 'households', 'latitude', 'longitude', 'housing\_median\_age', 'ocean\_proximity']

['median\_income', 'total\_rooms', 'total\_bedrooms'] -> ['median\_house\_value', 'population', 'households', 'latitude', 'longitude', 'housing\_median\_age', 'ocean\_proximity']

['median\_income', 'total\_rooms', 'households'] -> ['median\_house\_value', 'population', 'total\_bedrooms', 'latitude', 'longitude', 'housing\_median\_age', 'ocean\_proximity']

['median\_income', 'total\_rooms', 'latitude'] -> ['median\_house\_value', 'population', 'total\_bedrooms', 'households', 'longitude', 'housing\_median\_age', 'ocean\_proximity']

['median\_income', 'total\_rooms', 'longitude'] -> ['median\_house\_value', 'population', 'total\_bedrooms', 'households', 'latitude', 'housing\_median\_age', 'ocean\_proximity']

['median\_income', 'total\_rooms', 'housing\_median\_age'] -> ['median\_house\_value', 'population', 'total\_bedrooms', 'households', 'latitude', 'longitude', 'ocean\_proximity']

['median\_income', 'median\_house\_value', 'population'] -> ['ocean\_proximity']

['median\_income', 'median\_house\_value', 'population', 'latitude'] -> ['total\_rooms', 'total\_bedrooms', 'households', 'longitude', 'housing\_median\_age']

['median\_income', 'median\_house\_value', 'total\_bedrooms'] -> ['latitude', 'ocean\_proximity']

['median\_income', 'median\_house\_value', 'total\_bedrooms', 'households'] -> ['total\_rooms', 'population', 'longitude', 'housing\_median\_age']

['median\_income', 'housing\_median\_age', 'median\_house\_value', 'total\_bedrooms'] -> ['total\_rooms', 'population', 'households', 'longitude']

['median\_income', 'median\_house\_value', 'households'] -> ['ocean\_proximity']

['median\_income', 'housing\_median\_age', 'median\_house\_value', 'households'] -> ['total\_rooms', 'population', 'total\_bedrooms', 'latitude', 'longitude']

['median\_income', 'median\_house\_value', 'latitude', 'longitude'] -> ['housing\_median\_age']

['median\_income', 'population', 'total\_bedrooms'] -> ['total\_rooms', 'median\_house\_value', 'households', 'latitude', 'longitude', 'housing\_median\_age', 'ocean\_proximity']

['median\_income', 'population', 'households'] -> ['total\_rooms', 'median\_house\_value', 'total\_bedrooms', 'latitude', 'longitude', 'housing\_median\_age', 'ocean\_proximity']

['median\_income', 'population', 'latitude'] -> ['ocean\_proximity']

['median\_income', 'population', 'longitude'] -> ['total\_rooms', 'median\_house\_value', 'total\_bedrooms', 'households', 'latitude', 'housing\_median\_age', 'ocean\_proximity']

['median\_income', 'housing\_median\_age', 'population'] -> ['total\_rooms', 'median\_house\_value', 'total\_bedrooms', 'households', 'latitude', 'longitude', 'ocean\_proximity']

['median\_income', 'total\_bedrooms', 'households'] -> ['ocean\_proximity']

['median\_income', 'housing\_median\_age', 'total\_bedrooms', 'households'] -> ['total\_rooms', 'median\_house\_value', 'population', 'latitude', 'longitude']

['median\_income', 'total\_bedrooms', 'latitude'] -> ['ocean\_proximity']

['median\_income', 'housing\_median\_age', 'total\_bedrooms', 'latitude'] -> ['total\_rooms', 'median\_house\_value', 'population', 'households', 'longitude']

['median\_income', 'total\_bedrooms', 'longitude'] -> ['total\_rooms', 'median\_house\_value', 'population', 'households', 'latitude', 'housing\_median\_age', 'ocean\_proximity']

['median\_income', 'housing\_median\_age', 'total\_bedrooms'] -> ['ocean\_proximity']

['median\_income', 'households', 'latitude'] -> ['total\_rooms', 'median\_house\_value', 'population', 'total\_bedrooms', 'longitude', 'housing\_median\_age', 'ocean\_proximity']

['median\_income', 'households', 'longitude'] -> ['total\_rooms', 'median\_house\_value', 'population', 'total\_bedrooms', 'latitude', 'housing\_median\_age', 'ocean\_proximity']

['median\_income', 'housing\_median\_age', 'households', 'ocean\_proximity'] -> ['total\_rooms', 'median\_house\_value', 'population', 'total\_bedrooms', 'latitude', 'longitude']

['median\_income', 'housing\_median\_age', 'latitude', 'longitude'] -> ['median\_house\_value']

['median\_income', 'housing\_median\_age', 'latitude'] -> ['ocean\_proximity']

['median\_income', 'housing\_median\_age', 'longitude'] -> ['ocean\_proximity']

['total\_rooms', 'median\_house\_value', 'population'] -> ['median\_income', 'total\_bedrooms', 'households', 'latitude', 'longitude', 'housing\_median\_age', 'ocean\_proximity']

['total\_rooms', 'median\_house\_value', 'total\_bedrooms'] -> ['housing\_median\_age', 'ocean\_proximity']

['total\_rooms', 'median\_house\_value', 'households'] -> ['median\_income', 'population', 'total\_bedrooms', 'latitude', 'longitude', 'housing\_median\_age', 'ocean\_proximity']

['total\_rooms', 'median\_house\_value', 'latitude'] -> ['housing\_median\_age', 'ocean\_proximity']

['total\_rooms', 'median\_house\_value', 'longitude'] -> ['median\_income', 'population', 'total\_bedrooms', 'households', 'latitude', 'housing\_median\_age', 'ocean\_proximity']

['total\_rooms', 'population', 'total\_bedrooms'] -> ['median\_income', 'median\_house\_value', 'households', 'latitude', 'longitude', 'housing\_median\_age', 'ocean\_proximity']

['total\_rooms', 'population', 'households'] -> ['median\_income', 'median\_house\_value', 'total\_bedrooms', 'latitude', 'longitude', 'housing\_median\_age', 'ocean\_proximity']

['total\_rooms', 'population', 'latitude'] -> ['median\_income', 'median\_house\_value', 'total\_bedrooms', 'households', 'longitude', 'housing\_median\_age', 'ocean\_proximity']

['total\_rooms', 'population', 'longitude'] -> ['ocean\_proximity']

['housing\_median\_age', 'total\_rooms', 'population', 'longitude'] -> ['median\_income', 'median\_house\_value', 'total\_bedrooms', 'households', 'latitude']

['housing\_median\_age', 'total\_rooms', 'population', 'ocean\_proximity'] -> ['median\_income', 'median\_house\_value', 'total\_bedrooms', 'households', 'latitude', 'longitude']

['housing\_median\_age', 'total\_rooms', 'total\_bedrooms', 'households'] -> ['median\_income', 'median\_house\_value', 'population', 'latitude', 'longitude', 'ocean\_proximity']

['total\_rooms', 'total\_bedrooms', 'latitude'] -> ['median\_income', 'median\_house\_value', 'population', 'households', 'longitude', 'housing\_median\_age', 'ocean\_proximity']

['total\_rooms', 'total\_bedrooms', 'longitude'] -> ['median\_income', 'median\_house\_value', 'population', 'households', 'latitude', 'housing\_median\_age', 'ocean\_proximity']

['total\_rooms', 'households', 'latitude'] -> ['median\_income', 'median\_house\_value', 'population', 'total\_bedrooms', 'longitude', 'housing\_median\_age', 'ocean\_proximity']

['total\_rooms', 'households', 'longitude'] -> ['median\_income', 'median\_house\_value', 'population', 'total\_bedrooms', 'latitude', 'housing\_median\_age', 'ocean\_proximity']

['median\_house\_value', 'population', 'total\_bedrooms', 'households'] -> ['median\_income', 'total\_rooms', 'latitude', 'longitude', 'housing\_median\_age']

['housing\_median\_age', 'median\_house\_value', 'population', 'total\_bedrooms'] -> ['median\_income', 'total\_rooms', 'households', 'latitude', 'longitude', 'ocean\_proximity']

['ocean\_proximity', 'median\_house\_value', 'population', 'total\_bedrooms'] -> ['median\_income', 'total\_rooms', 'households', 'latitude', 'longitude', 'housing\_median\_age']

['median\_house\_value', 'population', 'households', 'latitude'] -> ['median\_income', 'total\_rooms', 'total\_bedrooms', 'longitude', 'housing\_median\_age']

['housing\_median\_age', 'median\_house\_value', 'population', 'households'] -> ['median\_income', 'total\_rooms', 'total\_bedrooms', 'latitude', 'longitude', 'ocean\_proximity']

['median\_house\_value', 'population', 'latitude'] -> ['ocean\_proximity']

['median\_house\_value', 'population', 'longitude'] -> ['median\_income', 'total\_rooms', 'total\_bedrooms', 'households', 'latitude', 'housing\_median\_age', 'ocean\_proximity']

['median\_house\_value', 'total\_bedrooms', 'households', 'latitude'] -> ['median\_income', 'total\_rooms', 'population', 'longitude', 'housing\_median\_age']

['median\_house\_value', 'total\_bedrooms', 'households', 'longitude'] -> ['median\_income', 'total\_rooms', 'population', 'latitude', 'housing\_median\_age']

['housing\_median\_age', 'median\_house\_value', 'total\_bedrooms', 'households'] -> ['median\_income', 'total\_rooms', 'population', 'latitude', 'longitude', 'ocean\_proximity']

['housing\_median\_age', 'median\_house\_value', 'total\_bedrooms', 'latitude'] -> ['ocean\_proximity']

['housing\_median\_age', 'median\_house\_value', 'total\_bedrooms', 'longitude'] -> ['median\_income', 'total\_rooms', 'population', 'households', 'latitude', 'ocean\_proximity']

['median\_house\_value', 'households', 'latitude'] -> ['ocean\_proximity']

['median\_house\_value', 'households', 'latitude', 'longitude'] -> ['median\_income', 'total\_rooms', 'population', 'total\_bedrooms', 'housing\_median\_age']

['median\_house\_value', 'households', 'longitude'] -> ['ocean\_proximity']

['housing\_median\_age', 'median\_house\_value', 'households', 'longitude'] -> ['median\_income', 'total\_rooms', 'population', 'total\_bedrooms', 'latitude']

['population', 'total\_bedrooms', 'households'] -> ['ocean\_proximity']

['housing\_median\_age', 'population', 'total\_bedrooms', 'households'] -> ['median\_income', 'total\_rooms', 'median\_house\_value', 'latitude', 'longitude']

['population', 'total\_bedrooms', 'latitude'] -> ['median\_income', 'total\_rooms', 'median\_house\_value', 'households', 'longitude', 'housing\_median\_age', 'ocean\_proximity']

['population', 'total\_bedrooms', 'longitude'] -> ['median\_income', 'total\_rooms', 'median\_house\_value', 'households', 'latitude', 'housing\_median\_age', 'ocean\_proximity']

['population', 'households', 'latitude'] -> ['ocean\_proximity']

['population', 'households', 'latitude', 'longitude'] -> ['median\_income', 'total\_rooms', 'median\_house\_value', 'total\_bedrooms', 'housing\_median\_age']

['housing\_median\_age', 'population', 'households', 'latitude'] -> ['median\_income', 'total\_rooms', 'median\_house\_value', 'total\_bedrooms', 'longitude']

['population', 'households', 'longitude'] -> ['ocean\_proximity']

['housing\_median\_age', 'population', 'households', 'longitude'] -> ['median\_income', 'total\_rooms', 'median\_house\_value', 'total\_bedrooms', 'latitude']

['housing\_median\_age', 'population', 'longitude'] -> ['ocean\_proximity']

['total\_bedrooms', 'households', 'latitude', 'longitude'] -> ['median\_income', 'total\_rooms', 'median\_house\_value', 'population', 'housing\_median\_age']

['housing\_median\_age', 'total\_bedrooms', 'households', 'latitude'] -> ['ocean\_proximity']

['housing\_median\_age', 'total\_bedrooms', 'households', 'longitude'] -> ['ocean\_proximity']

['housing\_median\_age', 'total\_bedrooms', 'latitude', 'longitude'] -> ['median\_income', 'total\_rooms', 'median\_house\_value', 'population', 'households']

['latitude', 'longitude'] -> ['ocean\_proximity']
\end{tcolorbox}

%% file: fds/loan.tex
\begin{tcolorbox}[
    colframe=gray!50!black,
    colback=gray!5!white,
    title=\small\texttt{Loan},
    breakable,
    enhanced jigsaw,
]
\scriptsize
\ttfamily
['Income', 'CITY'] -> ['House\_Ownership', 'Married/Single', 'Car\_Ownership']

['Income', 'CITY', 'Age'] -> ['Profession', 'Experience', 'CURRENT\_JOB\_YRS', 'CURRENT\_HOUSE\_YRS']

['Income', 'CITY', 'Profession'] -> ['Experience', 'CURRENT\_JOB\_YRS', 'CURRENT\_HOUSE\_YRS']

['Income', 'CITY', 'Experience'] -> ['Profession', 'CURRENT\_JOB\_YRS', 'CURRENT\_HOUSE\_YRS']

['CURRENT\_HOUSE\_YRS', 'Income', 'CITY'] -> ['Profession', 'Experience', 'CURRENT\_JOB\_YRS']

['Income', 'Age'] -> ['House\_Ownership', 'Married/Single', 'Car\_Ownership']

['Income', 'Age', 'Profession'] -> ['CITY', 'STATE', 'Experience', 'CURRENT\_JOB\_YRS', 'CURRENT\_HOUSE\_YRS']

['Income', 'Age', 'STATE'] -> ['CITY', 'Profession', 'Experience', 'CURRENT\_JOB\_YRS', 'CURRENT\_HOUSE\_YRS']

['Income', 'Age', 'Experience'] -> ['CITY', 'Profession', 'STATE', 'CURRENT\_JOB\_YRS', 'CURRENT\_HOUSE\_YRS']

['CURRENT\_HOUSE\_YRS', 'Income', 'Age'] -> ['CITY', 'Profession', 'STATE', 'Experience', 'CURRENT\_JOB\_YRS']

['Income', 'Age', 'Risk\_Flag'] -> ['CITY', 'Profession', 'STATE', 'Experience', 'CURRENT\_JOB\_YRS', 'CURRENT\_HOUSE\_YRS']

['Income', 'Profession'] -> ['House\_Ownership', 'Married/Single', 'Car\_Ownership']

['Income', 'Profession', 'STATE'] -> ['CITY', 'Experience', 'CURRENT\_JOB\_YRS', 'CURRENT\_HOUSE\_YRS']

['Income', 'Profession', 'Experience'] -> ['CITY', 'STATE', 'CURRENT\_JOB\_YRS', 'CURRENT\_HOUSE\_YRS']

['CURRENT\_HOUSE\_YRS', 'Income', 'Profession'] -> ['CITY', 'STATE', 'Experience', 'CURRENT\_JOB\_YRS']

['Income', 'STATE'] -> ['House\_Ownership', 'Car\_Ownership']

['Income', 'Married/Single', 'STATE', 'Experience'] -> ['Profession', 'CURRENT\_JOB\_YRS', 'CURRENT\_HOUSE\_YRS']

['CURRENT\_HOUSE\_YRS', 'Income', 'STATE'] -> ['CITY', 'Profession', 'Experience', 'CURRENT\_JOB\_YRS', 'Married/Single']

['Income', 'Married/Single', 'STATE'] -> ['CITY']

['Income', 'Experience'] -> ['House\_Ownership']

['CURRENT\_HOUSE\_YRS', 'Income', 'Experience'] -> ['CITY', 'Profession', 'STATE', 'CURRENT\_JOB\_YRS', 'Married/Single', 'Car\_Ownership']

['Income', 'CURRENT\_JOB\_YRS'] -> ['CITY', 'Profession', 'STATE', 'Experience', 'CURRENT\_HOUSE\_YRS', 'House\_Ownership', 'Married/Single', 'Car\_Ownership']

['CURRENT\_HOUSE\_YRS', 'Income', 'Car\_Ownership'] -> ['House\_Ownership']

['CITY'] -> ['STATE']

['CITY', 'Age', 'Profession', 'Experience', 'CURRENT\_JOB\_YRS'] -> ['House\_Ownership']

['CITY', 'Age', 'Profession', 'Experience', 'CURRENT\_JOB\_YRS', 'CURRENT\_HOUSE\_YRS'] -> ['Car\_Ownership']

['CITY', 'Age', 'Profession', 'Experience', 'CURRENT\_HOUSE\_YRS'] -> ['House\_Ownership', 'Married/Single']

['CITY', 'Age', 'Profession', 'Experience', 'CURRENT\_HOUSE\_YRS', 'Risk\_Flag', 'Car\_Ownership'] -> ['CURRENT\_JOB\_YRS']

['CITY', 'Age', 'Profession', 'Experience', 'Risk\_Flag', 'Married/Single'] -> ['House\_Ownership']

['Age', 'Profession', 'STATE', 'Experience', 'CURRENT\_JOB\_YRS', 'CURRENT\_HOUSE\_YRS', 'Risk\_Flag', 'Car\_Ownership'] -> ['Married/Single']
\end{tcolorbox}